\pdfoutput=1
\documentclass[11pt,dvipsnames]{article}

\usepackage[]{acl}

\usepackage{times}
\usepackage{latexsym}

\usepackage[T1]{fontenc}
\usepackage[utf8]{inputenc}

\usepackage{microtype}

\usepackage{colortbl}

\usepackage{xcolor}
\definecolor{thedarkblue}{RGB}{0,0,120} %104} % 180
\definecolor{mydarkblue}{rgb}{0,0.08,0.45} %ICML dark blue
\definecolor{darkblue}{rgb}{0,0.08,180}
\colorlet{TufteRed}{red!80!black}
\definecolor{theblue}{RGB}{0,0,180}
\colorlet{thered}{TufteRed}
      
\usepackage{microtype}
\usepackage{balance}

\usepackage{booktabs}
\usepackage{tabularx}

\usepackage{amsmath,amssymb,amsthm}

\newcommand{\eat}[1]{\ignorespaces}
\usepackage{comment}

\newcommand{\journal}[1]{} % text that is too long for short conf paper, but would be good for journal

\usepackage{tikz}
\usepackage{verbatim}
\usetikzlibrary{arrows}
\usetikzlibrary{shapes,snakes}
\usetikzlibrary{decorations.pathmorphing} % noisy shapes
\usetikzlibrary{fit}					% fitting shapes to coordinates
\usetikzlibrary{backgrounds}	% drawing the background after the foreground

\usepackage{ragged2e}
\usepackage{multirow}
\usepackage{microtype}
\usepackage{balance}
\usepackage{setspace}

\graphicspath{{./}{./graphics/}}
\newcolumntype{H}{>{\setbox0=\hbox\bgroup}c<{\egroup}@{}}

\newcolumntype{R}[1]{>{\RaggedLeft\arraybackslash}} %p{#1}}
\newcolumntype{L}[1]{>{\RaggedRight\arraybackslash}} %p{#1}}

\newcommand{\eg}{\emph{e.g.}}

\AtBeginEnvironment{pmatrix}{\setlength{\arraycolsep}{2pt}}

\providecommand{\mat}[1]{\boldsymbol{\mathrm{#1}}}%
\renewcommand{\vec}[1]{\boldsymbol{\mathrm{#1}}}

\DeclareMathOperator{\hugeE}{\mbox{\huge\raise-0.3ex\hbox{E}}}
\DeclareMathOperator{\p}{\mathbb{P}}
\DeclareMathOperator{\hugep}{\mbox{\huge\raise-0.3ex\hbox{$\p$}}}

\providecommand{\mX}{\ensuremath{\mat{X}}}

\providecommand{\vx}{\ensuremath{\vec{x}}}

\usepackage{subfigure}
\usepackage{float}
\usepackage{graphbox}
\usepackage{svg}
\usepackage{nicefrac}
\usepackage{xcolor}

\usepackage{bold-extra}
\usepackage[T1]{fontenc}

\usepackage{array}
\newcolumntype{P}[1]{>{\centering\arraybackslash}p{#1}}
\newcolumntype{M}[1]{>{\centering\arraybackslash}m{#1}}

\definecolor{orange}{rgb}{1,0.5,0}
\definecolor{graynode}{RGB}{20,20,20}
\definecolor{crimsonred}{RGB}{220,20,60}
\definecolor{darkgraynode}{gray}{0.5}
\definecolor{lightgraynode}{gray}{0.8}

\usepackage{rotate}
\usepackage{adjustbox}
\usepackage{array}
\usepackage{capt-of}
\usepackage{tabulary}
\usepackage{setspace}
\usepackage{amssymb}
\usepackage{mathtools}
\usepackage{pifont} 

\newcommand{\cmark}{\ding{51}}
\newcommand{\xmark}{\ding{55}}

\newcommand{\X}{\mathbb{X}}
\newcommand{\Y}{\mathbb{Y}}

\newcommand{\D}{\mathcal{D}}
\newcommand{\yhat}{\hat{Y}}

\definecolor{gray}{RGB}{20,20,20}
\definecolor{gray}{RGB}{0.7,0.7,0.7}
\definecolor{greencm}{RGB}{0,153,0}

\newcommand{\cm}{ {\color{greencm}\normalsize\cmark}}
\newcommand{\cmgray}{ {\color{gray}\normalsize\cmark}}
\newcommand{\xm}{ {\color{red}\normalsize\xmark}}

\newcommand{\hboldline}{\noalign{\hrule height 0.3mm}}
\newcommand{\boldbottomline}{\noalign{\hrule height 0.3mm}}

\definecolor{plotblue}{RGB}	{30,144,255}
\definecolor{plotgreen}{RGB}	{50,205,50}
\definecolor{plotred}{RGB}	{220,20,60}

\definecolor{myyellow}{RGB}{255,255,204}
\definecolor{myred}{RGB}{255,204,204}
\definecolor{myblue}{RGB}{0,200,255}
\definecolor{mygreen}{RGB}{80,220,80}

\newcommand*\hrulefillvar[1][0.4pt]{\leavevmode\leaders\hrule height#1\hfill\kern0pt}

\usepackage{algorithm}
\usepackage[noend]{algpseudocode}
\algrenewcommand\algorithmicrequire{\textbf{Input:}}
\algrenewcommand\algorithmicensure{\textbf{Output:}}

\usepackage{subfigure}
\usepackage{nicefrac}

\DeclareMathAlphabet{\mathbcal}{OMS}{cmsy}{b}{n}
\usepackage{amsmath}
\usepackage{mathrsfs}
\usepackage{comment}

\usepackage{bm}
\usepackage{bbm}
\usepackage{amssymb}

\usepackage[fixed]{fontawesome5}

\usepackage{paralist}
\usepackage{enumitem}

\definecolor{thedarkblue}{RGB}{0,0,120} %104} % 180
\definecolor{mydarkblue}{rgb}{0,0.08,0.45} %ICML dark blue

\usepackage{hyperref}
\hypersetup{%
    colorlinks=true,
    linkcolor=mydarkblue,
    citecolor=mydarkblue,
    filecolor=mydarkblue,
    urlcolor=mydarkblue}

\definecolor{googleblue}{HTML}{4285F4}
\definecolor{googlered}{HTML}{DB4437}
\definecolor{googlepurple}{HTML}{A142F4} % New purple color
\definecolor{googlegreen}{HTML}{0F9D58}

\definecolor{googleyellow}{HTML}{F4B400} % Bright yellow color 
\definecolor{googleorange}{HTML}{FBBC05} % Orange shade 
\definecolor{googlecyan}{HTML}{34A853} % Light cyan-green 
\definecolor{googlegray}{HTML}{9AA0A6} % Medium gray for text or subtle elements 
\definecolor{googlepink}{HTML}{EA4335} % Reddish-pink hue 
\definecolor{googlelightblue}{HTML}{7BAAF7} % Lighter blue for highlights or backgrounds

\usepackage{cleveref} %%added by subho

\usepackage{makecell}
\newcommand{\algcolor}[1]{\text{\textcolor{mydarkblue}{#1}}}

\newcommand{\writer}[1]{}

\newcommand{\hanjia}[1]{{\color{black}#1\color{black}}}
\newcommand{\todo}[1]{\textcolor{red}{~TODO:~#1}}

\newcommand{\ryan}[1]{\textcolor{blue}{~Ryan:~#1}}

\title{From Selection to Generation: A Survey of LLM-based Active Learning}
\author{
{\bf Yu Xia$^{1}$\thanks{~~Equal contributions}\hspace{1.5mm},
Subhojyoti Mukherjee$^{2\hspace{0.2mm}*}$,
Zhouhang Xie$^1$,
Junda Wu$^1$,
Xintong Li$^1$,} \\
{\bf
Ryan Aponte$^3$, Hanjia Lyu$^4$, Joe Barrow$^5$,
Hongjie Chen$^6$,
Franck Dernoncourt$^5$,
Branislav Kveton$^5$,} \\
{\bf
Tong Yu$^5$, Ruiyi Zhang$^5$, Jiuxiang Gu$^5$,
Nesreen K. Ahmed$^7$,
Yu Wang$^{8}$,
Xiang Chen$^5$,} \\
{\bf
Hanieh Deilamsalehy$^5$, Sungchul Kim$^{5}$,
Zhengmian Hu$^5$,
Yue Zhao$^9$, Nedim Lipka$^5$,
Seunghyun Yoon$^5$,} \\
{\bf Ting-Hao `Kenneth' Huang$^{10}$, Zichao Wang$^5$, Puneet Mathur$^5$, Soumyabrata Pal$^5$, Koyel Mukherjee$^5$,} \\
{\bf Zhehao Zhang$^{11}$, Namyong Park, Thien Huu Nguyen$^8$, Jiebo Luo$^4$, Ryan A. Rossi$^5$, Julian McAuley$^1$}\vspace{1mm}\\
$^1$University of California San Diego,
$^2$University of Wisconsin Madison, \\
$^3$Carnegie Mellon University,
$^4$University of Rochester,
$^5$Adobe Research,
$^6$Dolby Labs,\\
$^7$Cisco AI Research,
$^8$University of Oregon,
$^9$University of Southern California,\\
$^{10}$Pennsylvania State University,
$^{11}$Dartmouth College
}

\begin{document}
\maketitle

\begin{abstract}
Active Learning (AL) 
%with Large Language Models (LLMs) has emerged as 
has been a powerful paradigm for improving model efficiency and performance by selecting the most informative data points for labeling and training.
% With LLM-based AL, # RyanA- assuming this line typo
In recent active learning frameworks, Large Language Models (LLMs) have been employed not only for selection but also for generating entirely new data instances and providing more cost-effective annotations.
% \kenneth{As an opening sentence, it might be useful to mention the intuition: what's special about active learning *with LLM*?}\ryan{good point, added above sentence}
Motivated by the increasing importance of high-quality data and efficient model training in the era of LLMs, we present a comprehensive survey on LLM-based Active Learning.
%focusing on its use with LLMs, the various aspects that can be optimized,
% by LLM-based active learning, 
%its methods, applications, and integration with LLMs.
% for efficient data annotation. 
We introduce an intuitive taxonomy that categorizes these techniques and discuss the transformative roles LLMs can play in the active learning loop.
%how each category of techniques can be used for selecting various aspects for LLMs.
% We propose an intuitive taxonomy to classify the key approaches to AL in the context of LLMs, such as uncertainty sampling, query-by-committee, and diversity-based strategies, and explore their strengths and limitations. 
%Furthermore, we examine how LLMs are fine-tuned in AL settings, investigating both supervised and semi-supervised learning paradigms. 
We further examine the impact of AL on LLM learning paradigms and its applications across various domains. 
%various domains, such as natural language processing, question-answering, and dialogue systems. 
%Additionally, we discuss evaluation metrics and benchmark datasets commonly employed in the field, providing a comprehensive overview of their role in assessing AL performance with LLMs. 
Finally, we identify open challenges and propose future research directions.
%, such as model interpretability, scalability, and the integration of user feedback. 
This survey aims to serve as an up-to-date resource for researchers and practitioners seeking to gain an intuitive understanding of LLM-based AL techniques and deploy them to new applications.
% enhance the deployment of LLMs in active learning scenarios.
% This survey aims to serve as a valuable resource for researchers and practitioners seeking to understand and advance the development of personalized multimodal large language models.
\end{abstract}

\begin{figure}[t!]
\centering
\includegraphics[width=1.0\linewidth]{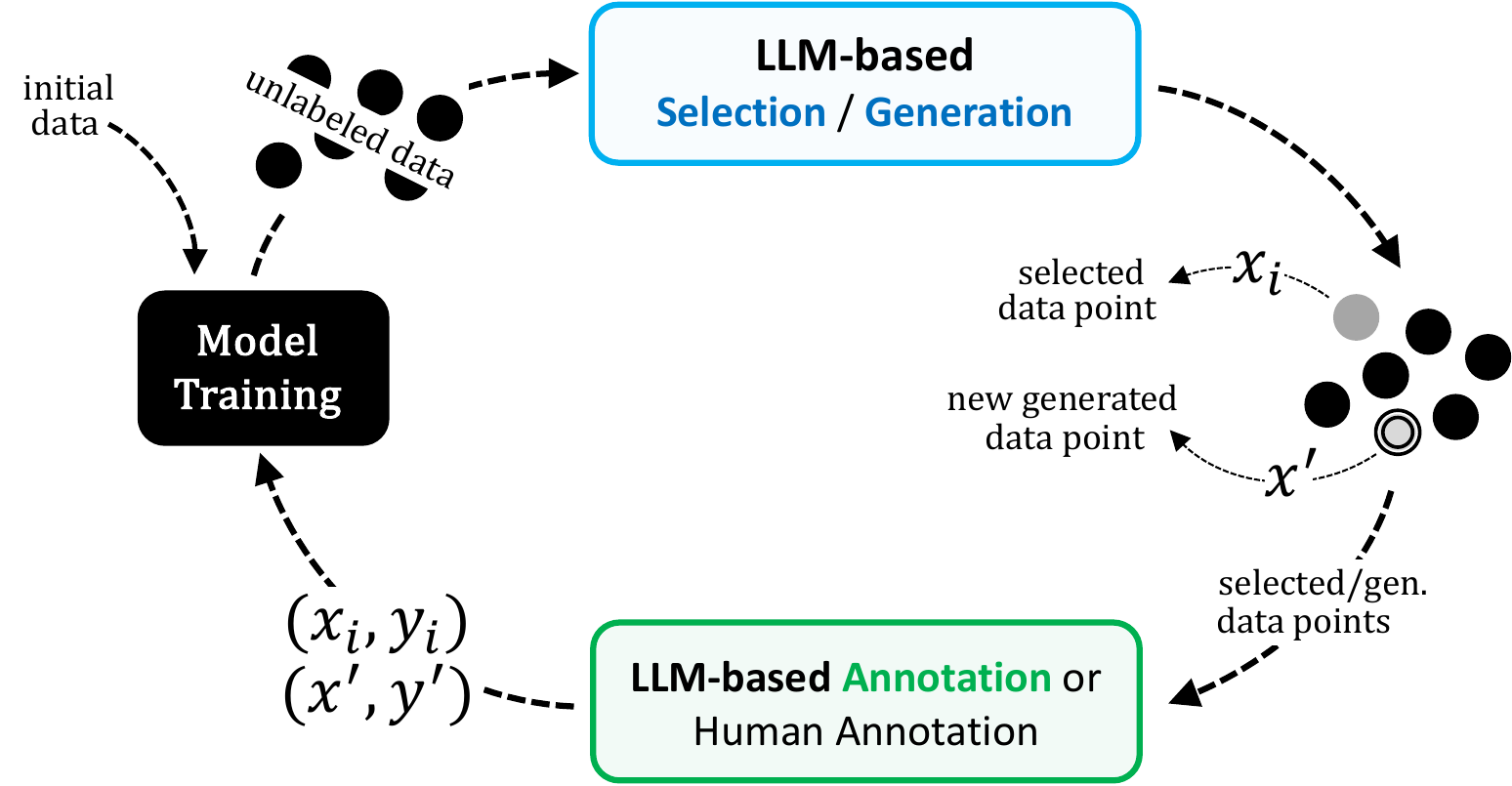}
\caption{Overview of LLM-based active learning. 
We start with initial data, including unlabeled instances $\mathcal{U}$.
There are two main steps.
First, LLM-based selection and/or generation leverages an LLM $\mathcal{M}$ to select unlabeled instances $\vx_i \in \mathcal{U}$ for annotation or \emph{generate entirely new instances} $\vx^{\prime} \not\in \mathcal{U}$. 
Next, given the LLM-based selected or generated instances $\vx_i$ or $\vx^{\prime}$, LLM-based annotation uses an LLM $\mathcal{M}$ to generate labels $y_i$ and $y^{\prime}$ for the instances.
Note that for intuition we show how LLMs can be leveraged for both steps; however, we may also use traditional techniques for selecting unlabeled instances, use humans for annotation, or both.
\eat{%}
Hence, an LLM-based AL approach is one that leverages LLMs for either of these steps, or both.
}%
% Note that in either step, one may also leverage 
% Note the above is an oversimplification
}
\label{fig:LLM-based-AL-overview}
%\vspace{4mm}
\end{figure}

\section{Introduction \writer{Ryan}}
Active Learning (AL) has been a widely studied technique that aims to reduce data annotation efforts by actively selecting most informative data samples for labeling and subsequent model training \cite{cohn1994improving,cohn1996active, settles2009active,olsson2009literature, fu2013survey,ren2021survey,zhan2022comparative}.
With an effective data selection strategy, this process helps to efficiently improve model performance with fewer labeled data instances, which can be particularly valuable when data annotation is expensive or time-consuming \cite{aggarwal2014active,hino2020active,schroder2020survey}.

\definecolor{googleblue}{HTML}{4285F4}
\definecolor{googlered}{HTML}{DB4437}
\definecolor{googlepurple}{HTML}{A142F4} % New purple color
\definecolor{googlegreen}{HTML}{0F9D58}

\begin{table*}[t!]
\centering
\small
\renewcommand{\arraystretch}{1.30} 
% \begin{tabular}{cr Xl}
\begin{tabularx}{1.0\linewidth}{
>{\centering\arraybackslash}m{18mm} 
>{\RaggedLeft\arraybackslash}m{44mm} 
X
% >{\arraybackslash}m{90mm} H
}
\toprule
\textbf{Class} & \textbf{General Mechanism} & 
\textbf{Description}
% \textbf{Example Models and Methods} 
\\ 
\midrule

\multirow{5}{*}{\textcolor{googlegreen}{\textbf{\makecell{Querying\\\textcolor{black}{(\textbf{Section~\ref{sec:querying}})}}}}} 
& Traditional Selection (Sec.~\ref{sec:querying-selection}) &  
This class of techniques uses traditional selection such as uncertainty sampling, disagreement, gradient-based sampling, and so on.
\\

& LLM-based Selection (Sec.~\ref{sec:querying-llm-based-selection}) &  
The class of LLM-based selection techniques focus on using LLMs to select the instances. 
% These techniques can be used 
\\

& LLM-based Generation (Sec.~\ref{sec:querying-llm-based-gen}) &  
The class of LLM-based generation techniques focus on generating novel instances.  \\

& Hybrid (Sec.~\ref{sec:querying-llm-based-selection-and-gen}) &  
Combines advantages of both LLM-based selection and generation
\\

\midrule

\multirow{5}{*}{\textcolor{googleblue}{\textbf{\makecell{Annotation\\\textcolor{black}{(\textbf{Section~\ref{sec:ann}})}}}}}
& Human Annotation (Sec.~\ref{sec:ann-human}) &  
Traditional human annotation simply refers to using humans to annotate the selected or generated instances, which is costly.
\\

& LLM-based Annotation (Sec.~\ref{sec:ann-llm-based}) &  
The class of LLM-based annotation techniques focus on leveraging LLMs for annotation and evaluation. This class of techniques are far cheaper than human annotation.
\\

& Hybrid (Sec.~\ref{sec:ann-hybrid-human-and-llm}) &  
This class of techniques aim to leverage the advantages of both humans and LLMs for optimal annotations while minimizing cost 
% (monetary, time, etc).
\\
% \midrule

% \multirow{2}{*}{\textcolor{googlegreen}{\textbf{\makecell{Hybrid\\\textcolor{black}{(\textbf{Section~\ref{sec:hybrid}})}}}}} &  TODO~\cite{} \\ 
% % \multirow{4}{*}{(\textbf{Section~\ref{sec:ret}})}
% & Alignment (Sec.~\ref{sec:ret-alignment}) &  TODO~\cite{} \\
% % & Generation (Sec.~\ref{sec:ret-generation}) & \citet{ye2024contemporary},Yo'LLaVA~\cite{nguyen2024yo} \\ 
% % & Fine-tuning (Sec.~\ref{sec:ret-finetune}) & FedPAM \cite{feng2024fedpam}, VITR \cite{gong2023vitr}\\ 

\bottomrule
\end{tabularx}
\caption{Taxonomy of LLM-based Active Learning Techniques (\Cref{sec:querying,sec:ann}).
% \ryan{%
% Active learning for LLMs vs. LLM for Active Learning
% }%
}
\label{tab:taxonomy-techniques}
%\vspace{-3mm}
\end{table*}

Despite the success of traditional active learning methods, the advent of Large Language Models (LLMs) with remarkable reasoning and generation capabilities creates a new paradigm of active learning.
For example, as illustrated in Figure~\ref{fig:LLM-based-AL-overview}, instead of solely relying on a predefined metric to query data instances, such as uncertainty \cite{wang2014new, diao2023active} or diversity \cite{agarwal2020contextual, citovsky2021batch}, LLMs can now be used to select most informative instances after reasoning or even generate entirely new instances that are better suited for efficient model training \cite{bayer2024activellm, parkar2024selectllm, yang2024rewards, bhatt2024experimental, zhang2024sieve}.
Moreover, with the collected informative data instances, LLMs also enable new data annotation schemes by collaborating with a human labeler or directly simulating a human labeler \cite{xiao2023freeal, kholodna2024llms, wang2024human}, which further reduces manual annotation efforts. 
LLM-based selection or generation can also help reduce training costs such as for supervised fine-tuning, in contrast to the main focus of traditional AL on reducing the labeling costs. 
%In contrast, the traditional AL focuses on reducing labeling cost, whereas in SFT, the goal is to reduce training cost, as there are no labels.

% \todob{Motivation may be different as well. Traditional AL focuses on labeling cost. In SFT, it is training cost because there are no labels.}

\begin{figure*}[t!]
\centering
\includegraphics[width=0.9\linewidth]{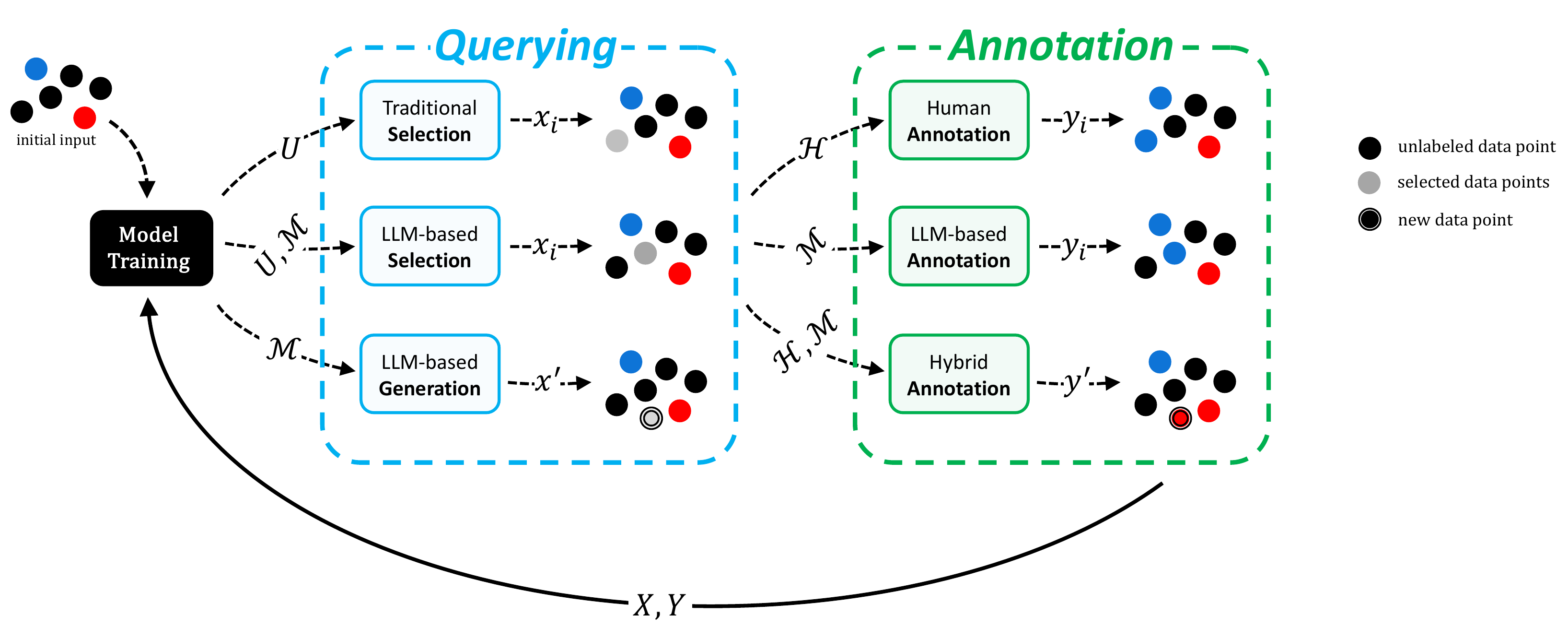}
\caption{%
Our proposed taxonomy classifies LLM-based active learning (AL) methods by their querying and annotation processes, the two key components of AL. 
Beyond selection,  LLMs enable the querying module to also generate unlabeled instances, while the annotation module assigns labels using LLMs, human annotators, or both.
% Overview of the proposed LLM-based active learning taxonomy.
}
\label{fig:LLM-based-AL}
%\vspace{4mm}
\end{figure*}

However, in spite of the immense potential of LLMs for active learning, particularly in high-quality data acquisition and annotation for efficient model training, existing surveys primarily focus on traditional active learning techniques, necessitating an up-to-date review of how LLMs have advanced AL in recent years. In this paper, we address this gap by presenting the first comprehensive survey of LLM-based AL techniques, which introduces a unifying taxonomy centered on the two main components of active learning: 
\textit{Querying} (selecting or generating unlabeled instances) and 
\textit{Annotation} (assigning labels).
\Cref{tab:taxonomy-techniques} and \Cref{fig:LLM-based-AL} provide an overview of the proposed taxonomy.
\Cref{table:qual-and-quant-comparison} further provide an intuitive comparison of existing LLM-based AL methods from the aspects of taxonomy and applications.
% Specifically, our survey systematically reviews relevant recent works as follows.
Guided by this taxonomy, our survey organizes and systematically reviews recent works across key aspects of LLM-based active learning as follows.
\begin{itemize}[nosep,leftmargin=0.8em]
	\item \textbf{Preliminaries} (\S\ref{sec:problem}):  We begin by introducing and formulating LLM-based active learning.
	\item \textbf{Querying} (\S\ref{sec:querying}): We describe querying strategies, including LLM-based selection and generation.
	\item \textbf{Annotation} (\S\ref{sec:ann}): We detail various annotation schemes, ranging from human annotation to LLM-based and hybrid approaches.
	\item \textbf{Stopping} (\S\ref{sec:termination}): 
%	We discuss how recent works incorporate LLM costs into the termination of the active learning loop.
	We discuss how recent works consider LLM costs for stopping the AL loop.
	\item \textbf{Active Learning Paradigms} (\S\ref{sec:setting}): 
	We examine how AL influences LLMs' learning paradigms.
	\item \textbf{Applications} (\S\ref{sec:apps}): We highlight the diverse applications of LLM-based active learning.
	\item \textbf{Open Problems} (\S\ref{sec:open-problems-challenges}): We discuss open problems and present future research directions.
\end{itemize}

\paragraph{Survey Scope}
This survey focuses mainly on recent works leveraging LLMs for AL, 
which creates a new paradigm driven by LLMs' reasoning and generation capabilities.
While some works use traditional AL for LLMs, 
we may not cover them thoroughly as they use techniques reviewed by prior surveys 
\cite{zhan2022comparative, perlitz2023active}.

\definecolor{googleblue}{RGB}{66,133,244}
\definecolor{googlered}{RGB}{219,68,55}
\definecolor{googlegreen}{RGB}{15,157,88}
\definecolor{googlepurple}{RGB}{138,43,226}
\definecolor{lightred}{RGB}{255, 220, 219}
\definecolor{lightblue}{RGB}{204, 243, 255}
\definecolor{lightgreen}{RGB}{200, 247, 200}
\definecolor{lightpurple}{RGB}{230,230,250}
\definecolor{lightyellow}{RGB}{242, 232, 99}
\definecolor{lighterblue}{RGB}{197, 220, 255}
\definecolor{lighterred}{RGB}{253, 249, 205}
\definecolor{lightyellow}{RGB}{207, 161, 13}
\definecolor{darkpurple}{RGB}{218, 210, 250}
\definecolor{darkred}{RGB}{255,198,196}
\definecolor{darkblue}{RGB}{172, 233, 252}

\section{What is LLM-based Active Learning? \writer{Joe Barrow}} \label{sec:problem}
%In this section, we introduce some preliminaries.
We start with basic notations and objective of traditional active learning and then introduce the LLM-based active learning loop with five main steps.

\paragraph{Traditional Active Learning} Let $\mathcal{U} = \{\vx_i\}_{i=1}^N$ be a pool of $N$ unlabeled instances, where $\vx_i \in \mathcal{X}$ are feature vectors in the input space $\mathcal{X}$. 
%\todob{This is confusing. Why would $\vx_i$ be a feature vector? It is a sequence of tokens. The labels $y_i$ below are also sequences of tokens.}
%However, we also have a potentially infinite set of unlabeled instances $\mathcal{U}_g$ that can be generated by an LLM $\mathcal{M}$.
Furthermore, let $\mathcal{L} = \{(\vx_i, y_i)\}_{i=1}^M$ be a labeled dataset, where $y_i \in \mathcal{Y}$ are the corresponding labels from the label space $\mathcal{Y}$ and the size of the labeled set $|\mathcal{L}| = M$ grows as more data instances are labeled.
We are also given an annotation budget $k$ that limits the number of instances that can be labeled by the annotator.
We have a target model $f_\theta : \mathcal{X} \to \mathcal{Y}$ parameterized by $\theta$ to be iteratively trained using the labeled data $\mathcal{L}$.
Note that the target model $f_\theta$ can be any parameterized predictive model, regardless of its architecture.
The objective of traditional AL is to efficiently select and label a subset of the unlabeled data instances $\vx_i \in \mathcal{U}$ to maximize the performance of model~$f_\theta$ before reaching the annotation budget $k$. 
%\todob{I am confused why we introduce both $k$ and $M$. In traditional active learning, we want to label $k \ll N$ examples $\vx_i \in \mathcal{U}$. Then, as we say later, one new trend is that the labeled examples may not belong to $\mathcal{U}$ when synthesized.}

%\todob{Simplify notation. The example can be unlabeled, $x \in \mathcal{U}$, or generated, $x \notin \mathcal{U}$. The label $y$ may exist, may be obtained from a human, or generated. Why do we need to subindex in $x_i \in \mathcal{U}$? Why do we want to distinguish between labels $y$ or $y'$? Look at \cref{alg:llm-based-AL} and follow the same notation.}

\paragraph{LLM-based Active Learning} LLM-based AL shares a similar goal. 
However, it is not bounded by the unlabeled set $\mathcal{U}$, but can also use an LLM $\mathcal{M}$ to generate entirely new data instances denoted as $\vx^{\prime} \not\in \mathcal{U}$ as well as generating corresponding labels $y^{\prime}$ by simulating a human annotator.
We define LLM $\mathcal{M}$ here as a decoder, encoder, or encoder-decoder architecture trained on a corpus of hundreds of millions to trillions of tokens following \citet{gallegos2024bias}.
We formulate in details the LLM-based AL loop as follows.

\begin{itemize}[left=0pt]
    \item \textbf{Initialize}: For a good starting point of the active learning loop, an LLM $\mathcal{M}$ can be used to annotate an initial set of labels or generate an initial dataset $\mathcal{L}_\text{init}$ to warm up the target model $f_\theta$. This approach overcomes the cold start problem that traditional AL methods face when there is no labeled data instance available and the initial model $f_\theta$ does not offer sufficient information for selecting informative data instances, especially when $f_\theta$ is not a pre-trained model.
    %\todob{LLMs are also pre-trained. So cold start is less of an issue.}
    \item \textbf{Query}: With an initialized model $f_\theta$, the \textbf{Querying} (\S\ref{sec:querying}) module is implemented to acquire the most informative data instances. Extending traditional AL methods that only select from the unlabeled set $\mathcal{U}$ with certain uncertainty or diversity metrics, the LLM $\mathcal{M}$ can now be used to select instances $\vx_i \in \mathcal{U}$ either by scoring or directly choosing, augment existing examples with generated paraphrases or contrast examples, or even generate entirely new instances $\vx^{\prime} \not\in \mathcal{U}$.
    \item \textbf{Annotate}: The acquired data instances $\vx_i$ or $\vx^{\prime}$ are then passed to the \textbf{Annotation} (\S\ref{sec:ann}) module to obtain corresponding labels $y_i$ or $y^{\prime}$. The LLM $\mathcal{M}$ can be used either in conjunction with, to augment, or even to entirely replace the human labeler. 
    Data instances $\vx_i \in \mathcal{U}$ that are selected from the unlabeled set are then excluded from $\mathcal{U}$. All labeled instances $(\vx_i, y_i)$ or $(\vx^{\prime}, y^{\prime})$ are then added to the labeled set $\mathcal{L}$.
    \item \textbf{Train}: With newly annotated data instances added to the labeled dataset $\mathcal{L}$, the target model $f_\theta$ is trained with a step update on its parameters $\theta$. 
    %with respect to the loss $\ell((\vx_i, y_i), {f_\theta})$. 
    %\todob{Drop $\ell((\vx_i, y_i), {f_\theta})$. We do not explain it properly and never use it again.}
    The updated model $f_{{\theta}}$ is then used to provide information for the querying module in the next iteration before a stopping criterion is met.
    \item \textbf{Stop}: The active learning loop reaches \textbf{Stopping} (\S\ref{sec:termination}) when a fixed annotation budget $k$ is reached, or when some property of the model $f_\theta$ being optimized, such as convergence, is satisfied.
    With LLMs, the budget can be not only based on the cost of human annotators, but also the cost of prompting the LLM or a combination of both.
\end{itemize}
We also summarize the LLM-based active learning loop in Algorithm \ref{alg:llm-based-AL}.
The goal of LLM-based AL in the general setting is to iteratively select or generate with an LLM most informative instances $\vx_i$ or $\vx^{\prime}$ for human or LLM labeling and subsequent model training. We discuss other variants of LLM-based AL goals in Appendix \ref{app:al_setting}.

\begin{algorithm}[!t]
\captionsetup{font=small}
    \caption{\; LLM-based Active Learning}
    % Framework}
	\label{alg:llm-based-AL}
	\begin{algorithmic}[1]
		\small
		\Require Unlabeled dataset $\mathcal{U}$,
        LLM $\mathcal{M}$,
        Annotation budget $k$
        
        % \!\!\!\!\!\!\!large language model $\mathcal{M}$
        
        %\!\!\!\!\!\!\!prompts $\mathcal{P}$
        
        % \!\!\!\!\!\!\!budget $k$

        % \Require Unlabeled data $\mathcal{U}$, large language model $\mathcal{M}$, 
        
        % budget $k$

        \smallskip
		\Ensure Trained model $f_\theta$, Labeled dataset $\mathcal{L}$
        \medskip
		
		\State $\mathcal{L}_\text{init},~\mathcal{U}~\leftarrow~\texttt{Initialize}(\mathcal{U},~\mathcal{M})$ 
        % \Comment{\algcolor{\textbf{Initialize/Seed} (\S\ref{sec:seed-initialize})}}

        \State $f_\theta~\leftarrow~\texttt{Train}(\mathcal{L}_\text{init})$ 
        % \Comment{\algcolor{\textbf{Model Training} (\S\ref{sec:model-training})}}
		
        \While{\textbf{not} \texttt{Stop}$(k, f_\theta, \mathcal{M})$}         \Comment{\algcolor{\textbf{Stopping} (\S\ref{sec:termination})}}
        % \Comment{\algcolor{\textbf{Stopping} (\S\ref{sec:terminate})}}
        
		\State $\vx~\leftarrow~\texttt{Query}(f_\theta,~\mathcal{U},~\mathcal{M})$ 
        \Comment{\algcolor{\textbf{Querying} (\S\ref{sec:querying})}}
		
        \State $(\vx, y)~\leftarrow~\texttt{Annotate}(\vx,~\mathcal{M})$ 
        \Comment{\algcolor{\textbf{Annotation} (\S\ref{sec:ann})}}
		
        \State \textbf{if } $\vx \in \mathcal{U}$ \textbf{ then } $\mathcal{U}~\leftarrow~\mathcal{U}\setminus\{\vx\}$
        
        \State $\mathcal{L}~\leftarrow~\mathcal{L} \cup \{(\vx, y)\}$
		
        \State $f_\theta~\leftarrow~\texttt{Train}(\mathcal{L})$ 
        % \Comment{\algcolor{\textbf{Updating Model} (\S\ref{sec:model-training})}}
		
        \EndWhile
        \State \textbf{return} $~f_{{\theta}^{*}},~\mathcal{L}$
	\end{algorithmic}
\end{algorithm}

\section{Query: From Selection to Generation \writer{Subho}} \label{sec:querying}
Active learning fundamentally seeks to maximize model performance with minimal annotation cost by carefully selecting most informative examples. 
Traditionally, this process has relied on uncertainty-based and diversity-based metrics \cite{settles2011theories, wang2014new, geifman2017deep, citovsky2021batch}. 
%\todob{Cite. You have the references in \cref{sec:querying-selection}.}
With the advent of LLMs, however, the paradigm is evolving from merely selecting examples from a fixed unlabeled pool to also generating new, high-value queries on demand, which can be especially helpful in reducing the training cost for supervised fine-tuning. 
%\todob{Also the training cost is sometimes the main motivation, such as in SFT.}

\subsection{Traditional Selection} \label{sec:querying-selection}

In this section, we briefly survey some of the key existing active learning query selection strategies. 
%All active learning query selection strategies try to balance uncertainty and diversity in selecting unlabeled examples for annotation. 
To effectively select informative examples, traditional methods capture how uncertain the model is in predicting the label of the example, i.e., uncertainty, or how different the chosen example is from the already selected examples in the labeled pool, i.e., diversity. 
%\todob{The uncertainty and diversity metrics come completely out of blue. We need to say first why the uncertainty and diversity matter.}
Uncertainty-based methods, such as \text{Least Confidence}~\citep{settles2009active, settles2011theories, wang2014new}, \text{Margin Sampling}~\citep{tong2001support, balcan2009agnostic, settles2009active}, and \text{Max-Entropy}~\citep{wang2014new, kremer2014active, diao2023active}, quantify how unsure a model is about its predictions. Complementary to these, diversity-based strategies—like \text{CoreSet}~\citep{sener2017active, geifman2017deep, citovsky2021batch} and \text{CDAL}~\citep{agarwal2020contextual}—ensure that selected examples cover varied regions of the input space. 
Optimal designs \citep{pukelsheim06optimal} are also a form of diversity based on information gain.
Hybrid approaches, including \text{BADGE}~\citep{ash2019deep, ash2021gone} and \text{BALD}~\citep{kirsch2019batchbald, pmlr-v70-gal17a}, strike a balance between these two aspects. For a comprehensive discussion of these traditional strategies, see \Cref{app:trad-strategy} and recent surveys in \citet{ren2021survey, li2024survey}. %\todob{Optimal designs \citep{pukelsheim06optimal} are a form of diversity based on information gain.}
%Some of the diversity-based selection strategies include \texttt{Coreset} \citep{sener2017active, geifman2017deep, citovsky2021batch}, \texttt{CDAL} \citep{agarwal2020contextual}. Some uncertainty-based methods include \texttt{Least confidence} \citep{settles2009active, settles2011theories, wang2014new}, \texttt{Margin} \citep{tong2001support, balcan2009agnostic, settles2009active}, \texttt{Max-entropy} \cite{wang2014new, kremer2014active, diao2023active}. Some methods try to trade-off between the two strategies like \texttt{Badge} \citep{ash2019deep, ash2021gone}, \texttt{Bald} \citep{kirsch2019batchbald, pmlr-v70-gal17a}. We discuss all these methods briefly in \Cref{app:trad-strategy}. Interested reader can read more recent works on deep active learning in \citet{ren2021survey, li2024survey}.

% \todos{This is just surveying traditional LLM selection strategies? I can write this up. 1. Coreset, 2. Least confidence, 3. Margin, 4. Max entropy, 5. Badge, 6. Bald. There are a few recent ones also based on optimal design I can briefly talk about that. There is also rejection sampling methods.}

% \subsubsection{Traditional Strategies}

\subsection{LLM-based Selection} \label{sec:querying-llm-based-selection}
% Selection via LLMs

With the emergence of LLMs, active learning strategies are being reimagined to exploit their powerful in-context reasoning and few-shot capabilities. 
In LLM-based selection, the model itself plays a dual role—both as a predictor and as a selector of informative queries. 
For instance, {ActiveLLM} \cite{bayer2024activellm} leverages an LLM to assess uncertainty and diversity in a completely unsupervised manner, making it particularly suitable in few-shot and model mismatch settings.
%
%bayer2024activellm, parkar2024selectllm
%\texttt{ActiveLLM}~\cite{bayer2024activellm} uses LLMs to select instances for the few-shot and model mismatch setting. Importantly, ActiveLLM can estimate uncertainty and diversity without any annotated data, and does not require training during the annotation process.
% 
%
Similarly, \text{ActivePrune} \cite{azeemi2024language} applies an LLM-driven approach to prune large unlabeled pools, reducing the computational burden of traditional acquisition functions for tasks such as translation, sentiment analysis, topic classification, and summarization.
%
%
%In contrast, \texttt{SelectLLM}~\citep{parkar2024selectllm} directly uses an LLM to select unlabeled examples by prompting LLMs. Specifically, \texttt{SelectLLM} uses LLMs to estimate
%the usefulness and impact of each example in the unlabeled pool and then refines these examples into subsets using k-NN to use them for few-shot learning with LLMs. 
In another line of work, \text{SelectLLM}~\citep{parkar2024selectllm} prompts LLMs directly to evaluate and rank the usefulness of unlabeled examples; the ranked instances are then refined using k-NN clustering to form effective few-shot demonstrations.
Ask-LLM \cite{sachdeva2024train} also directly prompts an instruction-tuned LLM to assess the quality of a training example.
Recent work by \citet{jeong2024llm} further demonstrates that LLMs can generate meaningful rankings of examples, which in turn can inform fine-tuning for downstream tasks. 
%\todob{I do not see Ask-LLM discussed.}
%\citet{jeong2024llm} show that LLMs can generate ranks of examples (based on their features), and then use the features of these ranked examples to fine-tune LLMs for downstream tasks.

% \todos{The above selection strategies roughly work for LLMs as well. There are two ways: If few shot/in-context learning then just pick 1-4, get their contextual representation from the LLM (I am considering an LLM that is more than a black box) and plug in the method to get the next most informative sample(s). If we also have the freedom to iteratively train the LLM with samples or batches of samples then use 5-6 as well. These require gradient computation information.}
% \ryan{LLMs are also used to select directly whatever it is we are doing active learning on}

\subsection{LLM-based Generation} \label{sec:querying-llm-based-gen}
Beyond selection from a fixed unlabeled set, LLMs also enable the generation of entirely new examples and labels, thereby extending the AL paradigm to an effectively infinite search space. 
Such extension is fundamentally different from an earlier concept \cite{angluin1988queries} that only queries memberships for hypothetical instances. 
In the following, we distinguish between generation strategies that remain within the confines of an existing unlabeled pool and those that extend beyond it.
%In this section, we discuss generating new examples using LLMs, and/or annotating these examples with LLMs and finally using them for few shot learning. 

% Generating new examples entirely

% Generating new label types

% Generating model

% \todos{This is actually the most interesting one and least studied somehow (I feel). We have to look around for this. I am guessing you are saying that generate both $x$ and $y$.}

% \ryan{I wrote below. This is one example where they actually do generation which is then used for selection, but there are many other works, and probably more so, on that actually generate new examples not in the "unlabeled set", and then label it, etc. This is probably more interesting (and what I focused on in the figures), at least to me, as it changes the AL paradigm and problem entirely. The AL problem is usually selecting from an unlabeled set, but LLMs make it that we are selecting from a possibly infinite set, since they can generate us new examples}

\paragraph{Generation for Selection within Unlabeled Set}
Several recent works integrate traditional selection metrics with LLM capabilities to improve query selection for few-shot learning. For example, \citet{margatina2023active} employ a combination of k-NN and perplexity-based strategies, demonstrating that uncertainty sampling generally underperforms in few-shot in-context learning settings with some evidence that this may change with larger models. In a similar vein, \citet{mukherjee2024experimental} highlight the effectiveness of experimental design techniques, such as G-Optimal design, for selecting high-impact unlabeled examples. Additionally, \citet{diao2023active} harness LLMs to generate multiple answers to a given question, using the variability among these answers as a proxy for uncertainty—albeit without venturing outside the initial dataset. Meanwhile, \text{EAGLE}~\citep{bansal2023large} first samples examples based on a conditional informativeness criterion and then leverages LLMs to generate in-context labels, streamlining the annotation process.

\paragraph{Generation for Selection outside Unlabeled Set}

A more radical departure from traditional AL involves generating new examples that are not present in the original unlabeled pool. The \text{APE} framework~\citep{qian2024ape} uses a Query-by-Committee strategy combined with chain-of-thought prompting to synthesize new prompts that are then sent to human annotators. Other works, such as those by \citet{yang2024rewards} and \citet{mukherjee2024multi}, generate both new examples and their labels using a trained LLM. These generated examples are then subjected to a rejection sampling process, ensuring that only those meeting predefined accuracy thresholds are retained. Similarly, \citet{yao2023beyond} use an explanation-generation model to produce human-guided rationales, which not only enhance label quality but also inform a novel diversity-based selection strategy akin to coreset sampling.

%A more radical departure from traditional AL involves generating new examples that are not present in the original unlabeled pool. 
%\text{APE}~\citep{qian2024ape} generates new prompts using a Query-by-Committee sampling strategy and chain-of-thought prompting style technique and then uses humans to label these prompts.
%
%\citet{yang2024rewards} and \cite{mukherjee2024multi} generate both new examples and their labels from a trained LLM, then uses rejection sampling to remove examples that fall below the expected threshold of accuracy for some objectives and then again finetune the LLM to align it better for multiple objectives simultaneously. 
%
%Similarly, \citet{yao2023beyond} implement an explanation generation model to produce explanations that are 
%guided by human explanations which are obtained using a diversity-based AL sampling strategy similar to coreset for obtaining the explanation annotations. Then they use a prediction
%model that utilizes those explanations for few shot predictions, and a novel data
%diversity-based AL sampling strategy similar to coreset for obtaining the explanation annotations.

% \ryan{An alternative way to organize this, and other sections is to focus on what LLMs are being used to actively optimize (models, data points, features, etc).}

\subsection{Hybrid} \label{sec:querying-llm-based-selection-and-gen}
Recognizing that neither pure selection nor generation can fully address all challenges in AL, recent work has begun to explore hybrid strategies that combine both. For instance, \texttt{NoiseAL} \citep{yuan2024hide} employs a two-stage process: small LLMs first identify promising unlabeled examples, which are then passed to an annotator LLM for labeling. Similarly, the Causal-guided Active Learning (CAL) framework \citep{du2024causal} integrates density-based clustering with LLM-driven query selection (e.g., via GPT-4) to autonomously identify and correct bias patterns in unlabeled data. Such hybrid methods seek to leverage the complementary strengths of LLMs and traditional human-in-the-loop strategies, ultimately pushing the boundaries of active learning toward more efficient and robust systems.
%In this section, we discuss hybrid LLM-based strategies for querying that leverage both LLM-based selection and LLM-based generation.
%
%For instance, \text{NoiseAL}~\citep{yuan2024hide} uses two small LLMs to select unlabelled examples and then uses another annotator LLMs to obtain labels of the samples. 
%Similarly, \text{CAL}~\citep{du2024causal} combines active learning with the causal mechanisms
%and proposes a casual-guided active learning
%(CAL) framework. They use a density-based clustering method and utilize LLMs like GPT4 to automatically and autonomously identify informative unlabeled samples and induce bias patterns in the examples.
%Such hybrid methods seek to leverage the complementary strengths of LLMs and traditional human-in-the-loop strategies, ultimately pushing the boundaries of active learning toward more efficient and robust systems.

% Joe Barrow to help
\section{Annotation: From Human to LLMs \writer{Xintong Li \& Hanjia}}
% LLM-based} 
\label{sec:ann}
Annotation has traditionally relied on human experts for high-quality labels. 
Recently, leveraging LLMs as annotators has further reduced annotation expenses, though challenges such as bias and label inconsistency. 
A hybrid approach that integrates human expertise with LLM-based annotation offers a promising solution, balancing efficiency and accuracy through dynamic task routing, verification mechanisms, and prompt engineering strategies.

\newcommand\TE{\rule{0pt}{2.0ex}}
\newcommand\BE{\rule[-1.1ex]{0pt}{0pt}}

{
\newcolumntype{C}{ >{\centering\arraybackslash} m{4cm} } 
\providecommand{\rotateDeg}{90}
\setlength{\tabcolsep}{1.8pt}
\providecommand{\rotDeg}{70}
\definecolor{verylightgreennew}{RGB}	{220,255,220}
\definecolor{verylightrednew}{RGB}		{255, 230, 230}
\definecolor{verylightreddarker}{HTML} {FFCBCB} 
\definecolor{verylightrednew}{RGB}		{255, 230, 230}
\definecolor{verylightrednewlighter}{RGB}		{255, 229, 239}
\definecolor{lightgraynew}{rgb}{0.95,0.95,0.95}
\definecolor{newgray}{RGB}{0.3,0.3,0.3}
\providecommand{\cellsz}{0.40cm} 
\providecommand{\cellszlg}{0.50cm} 
\providecommand{\cellszsm}{0.40cm} 
\renewcommand{\cm}{{\color{greencm}\normalsize\cmark}}
\renewcommand{\cmgray}{{\color{lightgraynew}\normalsize\cmark}}
\renewcommand{\xm}{{\color{verylightreddarker}\normalsize\xmark}}
\newcommand\BBBBB{\rule[1.6ex]{0pt}{1.6ex}}
\newcommand\BBBnew{\rule[-2.5ex]{0pt}{0pt}} 
\newcommand\BBBBBB{\rule[-1.1ex]{0pt}{0pt}} 
\newcommand{\sysName}[1]{{\sf
\BBBBBB
#1
}}
\providecommand{\cellsomewhat}{
\BBBBB
\cmgray
\cellcolor{verylightgreennew}
}
\providecommand{\cellno}{
\BBBBB
\xm
\cellcolor{verylightrednew}}
\providecommand{\cellyes}{
\BBBBB
\cm
\cellcolor{verylightgreennew}
}

\newcolumntype{H}{>{\setbox0=\hbox\bgroup}c<{\egroup}@{}}

\begin{table*}[t!]
\vspace{3mm}
\centering
\def\arraystretch{1.2} 
\scriptsize
% \footnotesize
% \small
\begin{minipage}{1.0\linewidth}
% \columnwidth}
{\begin{center}
\begin{tabular}
{P{2mm}
l
c 
!{\vrule width 0.8pt} 
% QUERYING
P{\cellszlg} P{\cellszlg} P{\cellszlg} 
P{\cellszlg} 
% @{}
!{\vrule width 0.6pt}
% ANNOTATION
P{\cellszlg} 
P{\cellszlg} 
P{\cellszlg}  
% @{}
!{\vrule width 0.6pt}
% APPLICATIONS
P{\cellszsm} 
P{\cellszsm} 
P{\cellszsm} 
P{\cellszsm} 
P{\cellszsm} 
P{\cellszsm} 
P{\cellszsm} 
P{\cellszsm} 
P{\cellszsm}
% P{\cellszsm} 
!{\vrule width 0.8pt}
@{}
}

& & 
& \multicolumn{4}{c!{\vrule width 0.6pt}}{\textcolor{googlegreen}{\textsc{\bfseries\scshape Querying}} 
% (\textbf{\S\ref{sec:querying}})
}
& \multicolumn{3}{c!{\vrule width 0.6pt}}{\textcolor{googleblue}{
\textsc{\bfseries\scshape Annotation}}
% (\textbf{\S\ref{sec:ann}})
% \textbf{\makecell{Annotation\\\textcolor{black}{(\textbf{Section~\ref{sec:ann}})}}}
}
& \multicolumn{9}{c!{\vrule width 0.6pt}}{\textcolor{googlered}{\textsc{\bfseries\scshape Applications}}} 
\\

& & 
\BBBnew
& \multicolumn{4}{c!{\vrule width 0.6pt}}{(\textbf{Section~\ref{sec:querying}})}
\BBBnew
& \multicolumn{3}{c!{\vrule width 0.6pt}}{(\textbf{Section~\ref{sec:ann}})}
& \multicolumn{9}{c!{\vrule width 0.6pt}}{(\textbf{Section~\ref{sec:apps}})} 
\\

% & 
% % \multicolumn{1}{l}{\textbf{\bfseries\scshape \small Method}} 
% & 
% & 
% % QUERYING
% \rotatebox{\rotateDeg}{\textbf{\S\ref{sec:querying-selection}~~Traditional Selection}} & 
% \rotatebox{\rotateDeg}{\textbf{\S\ref{sec:querying-llm-based-selection}~~LLM-based Selection}} &
% \rotatebox{\rotateDeg}{\textbf{LLM-based Generation (\S\ref{sec:querying-llm-based-gen})}} &
% \rotatebox{\rotateDeg}{\textbf{Hybrid (\S\ref{sec:querying-llm-based-selection-and-gen})}} &
% % ANNOTATION
% \rotatebox{\rotateDeg}{\textbf{Human Annotation (\S\ref{sec:ann-human})}} &
% \rotatebox{\rotateDeg}{\textbf{LLM-based Annotation (\S\ref{sec:ann-llm-based})}} &  
% \rotatebox{\rotateDeg}{\textbf{Hybrid (\S\ref{sec:ann-hybrid-human-and-llm})}} & 
% % ==========
% %  APPLICATIONS
% % ==========
% \rotatebox{\rotateDeg}{\textbf{Hybrid}} & 
% \rotatebox{\rotateDeg}{\textbf{Link prediction}} &
% \rotatebox{\rotateDeg}{\textbf{Link classification}} &
% \rotatebox{\rotateDeg}{\textbf{Node classification}} &
% \rotatebox{\rotateDeg}{\textbf{Clustering}} &
% \rotatebox{\rotateDeg}{\textbf{Other}} 
% \\
% \noalign{\hrule height 0.8pt}

& 
% \multicolumn{1}{l}{\textbf{\bfseries\scshape \small Method}} 
& 
& 
% QUERYING
\rotatebox{\rotateDeg}{\textbf{Traditional Selection (\S\ref{sec:querying-selection})}} & 
\rotatebox{\rotateDeg}{\textbf{LLM-based Selection (\S\ref{sec:querying-llm-based-selection})}} &
\rotatebox{\rotateDeg}{\textbf{LLM-based Generation (\S\ref{sec:querying-llm-based-gen})}} &
\rotatebox{\rotateDeg}{\textbf{Hybrid (\S\ref{sec:querying-llm-based-selection-and-gen})}} &
% ANNOTATION
\rotatebox{\rotateDeg}{\textbf{Human Annotation (\S\ref{sec:ann-human})}} &
\rotatebox{\rotateDeg}{\textbf{LLM-based Annotation (\S\ref{sec:ann-llm-based})}} &  
\rotatebox{\rotateDeg}{\textbf{Hybrid (\S\ref{sec:ann-hybrid-human-and-llm})}} & 
% ==========
%  APPLICATIONS
% ==========
\rotatebox{\rotateDeg}{\textbf{Text Classification}} & 
\rotatebox{\rotateDeg}{\textbf{Text Summarization}} &
\rotatebox{\rotateDeg}{\textbf{Classification}} &
\rotatebox{\rotateDeg}{\textbf{Question Answering}} &
\rotatebox{\rotateDeg}{\textbf{Entity Matching}} &
\rotatebox{\rotateDeg}{\textbf{Debiasing}} &
\rotatebox{\rotateDeg}{\textbf{Translation}} &
\rotatebox{\rotateDeg}{\textbf{Sentiment Analysis}} &
% translation, sentiment analysis
\rotatebox{\rotateDeg}{\textbf{Other}} 
\\
\noalign{\hrule height 0.8pt}

% An approach called APE~\cite{qian2024ape} focuses on an active prompt engineering approach for entity matching where at each iteration, a set of prompts are derived, and then evaluated by a committee of models, where the best is selected, and then the approach repeats.
& \sysName{$\mathsf{\sf \bf APE}$}~\cite{qian2024ape}
&
% QUERYING
& \cellno % Traditional Selection
& \cellno % LLM Selection
& \cellyes % LLM Gen
& \cellno % Hybrid
% ANNOTATION
& \cellno % human
& \cellyes % LLM
& \cellno % Hybrid
% APPLICATIONS
& \cellno 
& \cellno
& \cellno
& \cellno
& \cellyes 
& \cellno
& \cellno
& \cellno 
& \cellno 
\\
\hline

% \citet{li2024active} proposes LLM-Determined Curriculum Active Learning (LDCAL) that improves the stability of the active learner by selecting instances from easy to hard by using LLMs to determine the difficulty of a document.
& \sysName{$\mathsf{\sf \bf LDCAL}$}~\cite{li2024active}
&
% QUERYING
& \cellyes % Traditional Selection
& \cellyes % LLM Selection
& \cellno % LLM Gen
& \cellno % Hybrid
% ANNOTATION
& \cellyes % human
& \cellno % LLM
& \cellno % Hybrid
% APPLICATIONS
& \cellno 
& \cellyes 
& \cellno
& \cellno
& \cellno
& \cellno
& \cellno
& \cellno 
& \cellno 
\\
\hline

% ActiveLLM~\cite{bayer2024activellm} uses LLMs to select instances for the few-shot and model mismatch setting. Importantly, ActiveLLM can estimate uncertainty and diversity without any annotated data, and does not require training during the annotation process.
& \sysName{$\mathsf{\sf \bf ActiveLLM}$}~\cite{bayer2024activellm}
&
% QUERYING
& \cellno % Traditional Selection
& \cellyes % LLM Selection
& \cellno % LLM Gen
& \cellno % Hybrid
% ANNOTATION
& \cellno % human
& \cellno % LLM
& \cellno % Hybrid
% APPLICATIONS
& \cellyes 
& \cellno 
& \cellno
& \cellno
& \cellno
& \cellno
& \cellno
& \cellno 
& \cellno 
\\
\hline

% ActivePrune~\cite{azeemi2024language} introduces a language model approach for pruning unlabeled instances for active learning settings where the unlabeled instance pool is large, and thus, computationally costly for an acquisition function to search over.
% Experiments on translation, sentiment analysis, topic classification, and summarization tasks
& \sysName{$\mathsf{\sf \bf ActivePrune}$}~\cite{azeemi2024language}
&
% QUERYING
& \cellyes % Traditional Selection
& \cellyes % LLM Selection
& \cellno % LLM Gen
& \cellno % Hybrid
% ANNOTATION
& \cellyes % human
& \cellno % LLM
& \cellno % Hybrid
% APPLICATIONS
& \cellyes 
& \cellyes 
& \cellno
& \cellno
& \cellno
& \cellno
& \cellyes
& \cellyes 
& \cellno 
\\
\hline

% \citet{ming2024autolabel} proposed AutoLabel
& \sysName{$\mathsf{\sf \bf AutoLabel}$}~\cite{ming2024autolabel}
&
% QUERYING
& \cellyes % Traditional Selection
& \cellno % LLM Selection
& \cellno % LLM Gen
& \cellno % Hybrid
% ANNOTATION
& \cellno % human
& \cellno % LLM
& \cellyes % Hybrid
% APPLICATIONS
& \cellno 
& \cellno 
& \cellno
& \cellno
& \cellyes
& \cellno
& \cellno
& \cellno 
& \cellno 
\\
\hline

% \citet{zhang2023llmaaa} proposed LLMaAA that leverages LLMs for annotation in an active learning loop. LLMaAA is used for both named entity recognition and relation extraction.
& \sysName{$\mathsf{\sf \bf LLMaAA}$}~\cite{zhang2023llmaaa}
&
% QUERYING
& \cellyes % Traditional Selection
& \cellno % LLM Selection
& \cellno % LLM Gen
& \cellno % Hybrid
% ANNOTATION
& \cellno % human
& \cellyes % LLM
& \cellno % Hybrid
% APPLICATIONS
& \cellno 
& \cellno 
& \cellno
& \cellno
& \cellyes
& \cellno
& \cellno
& \cellno 
& \cellyes % Other (RELATION EXTRACTION)
\\
\hline

% \cite{diao2023active} uses LLMs to generate $k$ answers to a question that are then used to measure the uncertainty for selection.
% In this work, they simply leverage LLMs to generate answers which are used to estimate the uncertainty of a given question, however, they do not generate new questions not in the initial dataset.
& \sysName{$\mathsf{\sf \bf Active\text{-}Prompt}$}~\cite{diao2023active}
&
% QUERYING
& \cellno % Traditional Selection
& \cellno % LLM Selection
& \cellyes % LLM Gen
& \cellno % Hybrid
% ANNOTATION
& \cellyes % human
& \cellno % LLM
& \cellno % Hybrid
% APPLICATIONS
& \cellno 
& \cellno 
& \cellno
& \cellyes
& \cellno
& \cellno
& \cellno
& \cellno 
& \cellno % Other (RELATION EXTRACTION)
\\
\hline

% \cite{rouzegar2024enhancing} investigates using LLMs with human annotations for text classification to achieve lower costs while maintaining accuracy. 
& \sysName{$\mathsf{\sf \bf HybridAL}$}~\cite{rouzegar2024enhancing}
&
% QUERYING
& \cellyes % Traditional Selection
& \cellno % LLM Selection
& \cellno % LLM Gen
& \cellno % Hybrid
% ANNOTATION
& \cellno % human
& \cellno % LLM
& \cellyes % Hybrid
% APPLICATIONS
& \cellyes 
& \cellno 
& \cellno
& \cellno
& \cellno
& \cellno
& \cellno
& \cellno 
& \cellno % Other 
\\
\hline

% ACL 2024
% https://aclanthology.org/2024.acl-long.592.pdf
& \sysName{$\mathsf{\sf \bf NoiseAL}$}~\cite{yuan2024hide} & 
% QUERYING
& \cellno % Traditional Selection
& \cellno % LLM Selection
& \cellyes % LLM Gen
& \cellyes % Hybrid
% ANNOTATION
& \cellno % human
& \cellyes % LLM
& \cellno % Hybrid
% APPLICATIONS
& \cellyes 
& \cellno 
& \cellno
& \cellno
& \cellno
& \cellno
& \cellno
& \cellno 
& \cellyes 
\\
\hline

% ACL 2024
% https://aclanthology.org/2024.acl-long.778.pdf
& \sysName{$\mathsf{\sf \bf CAL}$}~\cite{du2024causal} & 
% QUERYING
& \cellno % Traditional Selection
& \cellno % LLM Selection
& \cellno % LLM Gen
& \cellyes % Hybrid
% ANNOTATION
& \cellyes % human
& \cellno % LLM
& \cellno % Hybrid
% APPLICATIONS
& \cellno 
& \cellno 
& \cellno
& \cellno
& \cellno
& \cellyes
& \cellno
& \cellno 
& \cellno 
\\
\hline

% ICML 2024
% https://openreview.net/pdf?id=CTgEV6qgUy
& \sysName{$\mathsf{\sf \bf APL}$}~\cite{muldrew2024active} & 
% QUERYING
& \cellyes % Traditional Selection
& \cellno % LLM Selection
& \cellno % LLM Gen
& \cellno % Hybrid
% ANNOTATION
& \cellyes % human
& \cellno % LLM
& \cellno % Hybrid
% APPLICATIONS
& \cellno 
& \cellyes 
& \cellno
& \cellno
& \cellno
& \cellno
& \cellno
& \cellno 
& \cellyes
\\
\hline

% PKDD 2024
% https://arxiv.org/pdf/2404.02261
& \sysName{$\mathsf{\sf \bf AL\text{-}Loop}$}~\cite{kholodna2024llms} & 
% QUERYING
& \cellyes % Traditional Selection
& \cellno % LLM Selection
& \cellno % LLM Gen
& \cellno % Hybrid
% ANNOTATION
& \cellno % human
& \cellyes % LLM
& \cellno % Hybrid
% APPLICATIONS
& \cellno 
& \cellno 
& \cellno
& \cellno
& \cellyes
& \cellno
& \cellno
& \cellno 
& \cellno 
\\
\hline

% Neurips 2024
% https://arxiv.org/pdf/2406.10023
& \sysName{$\mathsf{\sf \bf BAL\text{-}PM}$}~\cite{melo2024deep} & 
% QUERYING
& \cellno % Traditional Selection
& \cellyes % LLM Selection
& \cellno % LLM Gen
& \cellno % Hybrid
% ANNOTATION
& \cellyes % human
& \cellno % LLM
& \cellno % Hybrid
% APPLICATIONS
& \cellno 
& \cellyes 
& \cellno
& \cellno
& \cellno
& \cellno
& \cellno
& \cellno 
& \cellno 
\\
\hline

% EMNLP 2023
% https://arxiv.org/pdf/2311.15614
& \sysName{$\mathsf{\sf \bf FreeAL}$}~\cite{xiao2023freeal} & 
% QUERYING
& \cellno % Traditional Selection
& \cellno % LLM Selection
& \cellno % LLM Gen
& \cellyes % Hybrid
% ANNOTATION
& \cellno % human
& \cellyes % LLM
& \cellno % Hybrid
% APPLICATIONS
& \cellyes 
& \cellno 
& \cellno
& \cellno
& \cellyes
& \cellno
& \cellno
& \cellyes 
& \cellno \\
\hline

% EMNLP Finding 2023 
% https://arxiv.org/pdf/2305.14264
& \sysName{$\mathsf{\sf \bf AL\text{-}Principle}$}~\cite{margatina2023active} & 
% QUERYING
& \cellyes % Traditional Selection
& \cellno % LLM Selection
& \cellno % LLM Gen
& \cellno % Hybrid
% ANNOTATION
& \cellyes % human
& \cellno % LLM
& \cellno % Hybrid
% APPLICATIONS
& \cellyes 
& \cellno 
& \cellyes
& \cellyes  % multiple-choice tasks
& \cellno
& \cellno
& \cellno
& \cellyes 
& \cellno 
\\
\hline

% EMNLP Finding 2023 
% https://arxiv.org/pdf/2305.12710

& \sysName{$\mathsf{\sf \bf Beyond\text{-}Labels}$}~\cite{yao2023beyond}
&
% QUERYING
& \cellno % Traditional Selection
& \cellno % LLM Selection
& \cellyes  % LLM Gen
& \cellno % Hybrid
% ANNOTATION
& \cellyes % human
& \cellno % LLM
& \cellno % Hybrid
% APPLICATIONS
& \cellyes 
& \cellno 
& \cellno
& \cellno
& \cellno
& \cellno
& \cellno
& \cellno 
& \cellyes 
\\
\hline

% Other potentially related works
% CVPR 2024 Active Prompt Learning in Vision Language Models
% https://ieeexplore.ieee.org/document/10655710

% arxiv 2023 Large Language Models as Annotators: Enhancing Generalization of NLP
% https://arxiv.org/pdf/2306.15766

% & \sysName{$\mathsf{\sf \bf TODO}$}~\cite{todo}
% &
% % QUERYING
% & \cellno % Traditional Selection
% & \cellno % LLM Selection
% & \cellno % LLM Gen
% & \cellno % Hybrid
% % ANNOTATION
% & \cellno % human
% & \cellno % LLM
% & \cellno % Hybrid
% % APPLICATIONS
% & \cellno 
% & \cellno 
% & \cellno
% & \cellno
% & \cellno
% & \cellno
% & \cellno
% & \cellno 
% & \cellno 
% \\
% \hline

\noalign{\hrule height 0.7pt}
\end{tabular}
\end{center}
}

\end{minipage}
\vspace{-1mm}
\caption{%
Overview of the proposed taxonomy for LLM-based active learning techniques and their applications.
% Techniques are categorized by the type of input graph supported, whether the approach focuses on pre-processing, in-processing (in-training), or post-processing, as well as the task the technique was designed.
Using this taxonomy, we provide a qualitative and quantitative comparison of LLM-based active learning methods.
% \hongjie{
% There are 3 rows I want someone to verify and confirm the taxonomy is accurate.
% CAL: this work utilizes AL to debias LLMs. The querying stage seems to depend on LLMs, but there is no annotation stage. 
% APL: not using LLMs for querying or annotation, but use it to rank selected samples; 
% AL-Principles: use LLM models for evaluation; Beyond-labels: Please verify LLM Generation is accurate, it is used for explanation generation, but not sample generation.
% }
% \namyong{
% I'll check if the check marks for the three papers (CAL, AL-Principle, Beyond-Labels) are accurate.
% }
% \namyong{
% Marked LDCAL as (1) using both traditional and LLM-based selection, and as (2) using human annotation. In LDCAL, LLMs are used for curriculum learning (evaluating the difficulty of training instances), but not for selection. However, I suppose its use for curriculum learning can be seen as an indirect form of selection.
% }
}
\label{table:qual-and-quant-comparison}
% \vspace{-3.5mm}
% \vspace{-6mm}
%\vspace{-3mm}
\end{table*}
}

\subsection{Human Annotation \writer{Xintong Li \& Hanjia}} \label{sec:ann-human}
Traditional human annotation involves sending selected or generated instances to annotators for labeling, which remains the most accurate approach but is often costly. Several recent works have explored active learning strategies to optimize the annotation process. 
ActivePrune~\cite{azeemi2024language} and CAL~\cite{du2024causal} reduce annotation costs by actively selecting the most valuable instances for labeling. 
\hanjia{For instance, in the case of imbalanced data, enhancing the model's performance on the minority class can not only improve overall accuracy but also mitigate biases. One effective approach to achieving this is by increasing the size of the minority class. Providing human annotators with data that includes more minority samples has been shown to be effective~\cite{lyu2022social}. }
Active-Prompt~\cite{diao2023active} and AL-Principle~\cite{margatina2023active} focus on guiding LLMs by selecting instances where annotators verify final answers or prompt examples before model predictions. 
Furthermore, Beyond-Labels~\cite{yao2023beyond} extends traditional annotation by collecting short rationales or natural language explanations alongside labels, improving both model interpretability and performance. 
Additionally, APL~\cite{muldrew2024active} and BAL-PM~\cite{melo2024deep} incorporate human preference learning by asking annotators to compare or rank multiple model outputs. 
\hanjia{However, there are still several challenges in human annotation. Human annotator variability, as differences in expertise, cognitive biases, and annotation consistency can lead to label noise and disagreements, ultimately affecting model performance. Moreover, bias and fairness considerations in active learning are also an active area of research, as human annotations can inadvertently reflect societal biases, leading to skewed model predictions.} 
{There are also works on active learning in RLHF and DPO, which we refer to Section \ref{app:setting} for more details.}

\subsection{LLM-based Annotation \writer{Hanjia}} \label{sec:ann-llm-based}
\hanjia{To address the challenges in human annotation, ongoing research explores leveraging LLMs as annotators in active learning, primarily to reduce annotation costs. 
FreeAL~\cite{xiao2023freeal}, which distills task-specific knowledge with the help of a downsteam small language model, demonstrates improved performance without any human supervision.
Similarly, by employing {GPT-4-Turbo} for annotating low-resource languages, \citet{kholodna2024llms} reported substantial reductions in estimated annotation costs compared to human annotation.
However, a key challenge in LLM-based annotation is ensuring high-quality labels. To mitigate quality issues, LLMaAA~\cite{zhang2023llmaaa} incorporates in-context examples, demonstrating improved annotation reliability.
Despite such advancements, LLMs, like human annotators, are susceptible to biases inherited from their training data. Research has shown that LLMs' responses to cognitive psychological tasks often resemble those of individuals from western, educated, industrialized, rich, and democratic societies~\cite{atari2023humans}. Biases in LLM-based annotations also persist in certain domains, such as political science~\cite{zhang2024electionsim}. For example, \citet{qi2024representation} identified three dimensions of bias in LLM-generated political samples: societal and cultural contexts, demographic groups, and political institutions.
A particular concerning case arises when LLMs are used to annotate content that has also been generated by LLMs. In such scenarios, there is a risk of \textit{self-reinforcement bias}, where the model's inherent tendencies are amplified rather than corrected. This could create a feedback loop where the model's performance may appear artificially inflated.
LLM-based annotations can also be sensitive to input variations~\cite{mizrahi-etal-2024-state}. Slight changes in prompt phrasing, context, or model sampling parameters can lead to inconsistent annotations~\cite{10.1145/3689217.3690621}. Ensuring consistency in LLM-generated annotations remains an ongoing challenge, requiring further research into prompt engineering, calibration techniques, and hybrid human-LLM validation strategies.
}
\begin{table*}[!t]
\centering
\small
\begin{tabular}{rlH}
\toprule
\multicolumn{1}{c}{\textbf{Task}} & \textbf{Description} & \textbf{Use Case} \\ 
\midrule
\textbf{Text Classification } (Sec.\ref{apps-text-classification}) & Selects uncertain or ambiguous texts for classification. & Sentiment analysis, topic categorization. \\ 
% \midrule

\textbf{Text Summarization } (Sec.\ref{apps-text-summarization}) & Chooses diverse or complex document types to improve summarization. & Summarizing news articles, scientific papers, etc. \\

\textbf{Non-Text Classification  } (Sec.\ref{apps-non-text-classification}) & The pairing of non-text samples, such as images, with labels. & TODO. \\ 

\textbf{Question Answering }  (Sec.\ref{apps-question-answering})  & Chooses ambiguous or difficult questions for QA systems to refine. & Answering questions in specialized domains like law or medicine. \\ 

\textbf{Entity Modeling }  (Sec.\ref{apps-entity-matching})  & A binary classification task to pair entities with one of two labels. & TODO. \\ 

\textbf{Debiasing }  (Sec.\ref{apps-active-debiasing})  & The process of reducing measured bias in machine-generated output. & TODO. \\ 

\textbf{Translation } (Sec.\ref{apps-translation})  & Selects sentences or phrases with high uncertainty in translation. & Improving accuracy of translations, especially for low-resource languages. \\ 

\textbf{Sentiment Analysis } (Sec.\ref{apps-sentiment-analysis})  & Focuses on sentences where the model is uncertain about sentiment. & Determining sentiment in reviews, social media posts, etc. \\ \bottomrule

% \midrule
\end{tabular}
\caption{
% \todo{Ryan A. to update, make consistent with subsections in application section, and add references for each} 
% \todo{Ryan Aponte: please also reorder subsections, to be same order as in Table 2, etc.}
Applications of LLM-based Active Learning.
%\ryan{This table needs to highlight the unique aspects for each application offered by LLM AL}
}
\label{tab:applications}
\end{table*}

\subsection{Hybrid \writer{Hanjia \& Xintong}} \label{sec:ann-hybrid-human-and-llm}
% In this section, we discuss techniques that leverage both humans and LLMs to obtain better annotations.
\hanjia{While LLM-based annotation has significantly reduced costs, it remains prone to errors, particularly in complex or domain-specific tasks~\cite{lu2023human}. To address this issue, researchers have developed methods to evaluate annotation quality and dynamically route data to either LLMs or human annotators, balancing efficiency and accuracy. For example, \citet{wang2024human} proposed a multi-step human-LLM collaborative approach where LLMs first generate labels and explanations. A verifier then assesses the quality of these labels, and human annotators re-annotate a subset of low-quality cases. Similarly, }
\citet{rouzegar2024enhancing} investigates using LLMs with human annotations for text classification to achieve lower costs while maintaining accuracy \hanjia{based on confidence thresholds}. 
\hanjia{Another approach to combining human expertise with LLMs involves having humans curate a set of annotated examples, which are then incorporated into LLM prompts to enable annotation in a few-shot learning manner. While this method is both intuitive and effective, selecting optimal examples remains challenging. \citet{qiu2024semantics} show that examples included in prompts can sometimes overly constrain LLM decision-making, leading models to favor labels that align closely with provided examples rather than considering a broader range of possibilities. Refining strategies for selecting and structuring prompt examples is therefore an important direction for improving LLM-assisted annotation.}

\section{Stopping: From Criterion to LLMs \writer{Junda}} \label{sec:termination}
Stopping criterion in active learning loop is crucial for balancing model performance improvements with annotation costs. 
In LLM-based AL, stopping criteria must consider not only traditional factors such as annotation budget $k$, model performance gains or uncertainty reduction, 
but also the variable costs associated with LLMs.

%\todob{I do not understand this section. When we formulated the problem, we clearly said that the stopping criterion is the budget $k$. Why do we need other termination criteria?}

\subsection{Traditional Approaches} \label{sec:termination-traditional}
In traditional active learning, 
one widely adopted strategy is to stop querying when performance improvements on a validation set fall below a predefined threshold \cite{settles2009active}.
Other approaches monitor the stability of the model’s predictions or the uncertainty estimates, terminating annotation once these metrics indicate that the model has sufficiently converged \cite{tong2001active}.
In addition, several theoretical frameworks provide sample complexity bounds as termination criteria, ensuring that marginal returns are decreasing \cite{zhu2003combining,hanneke2007bound,dasgupta2009analysis}.
However, such criteria assume that the cost per annotation is uniform and homogeneous, which simplifies the stopping rule.

\subsection{Cost-Aware Termination} \label{sec:termination-hybrid}
In LLM-based AL, estimating the annotation cost can be challenging and complex.
While traditional AL relies on a discrete budget $k$ to represent the number of human annotations, 
LLM-based AL may combine both human and LLM annotations. 
In such cases, the cost of an annotation depends not only on the source (human vs. LLM) but also on variable factors like input and output token counts.
Even in the case that LLM-based annotations are used without any human annotations, the cost cannot be easily approximated as the discrete budget that often represents the amount of examples that can be labeled.
It is straightforward to see that since the cost depends on the input and more so the output tokens, then the budget may be better represented as a real-valued amount that pertains directly to a monetary cost.
Recent exploration in hybrid stopping criteria \cite{akins2024cost,pullar2024hitting} including cost-aware termination criterion that integrates token-level cost analysis and performance plateau detection. 
Other works have suggested a combined cost-performance metric that balances the fixed costs of human annotations with the variable costs of LLM-based annotations  \cite{xia2024llm, zhang2024interactive}.

% amount of  may be a real-valued is no longer discrete, nor can be easily approximated with the budget $b$, as $b$ in both cases 

% annotating making the cost 
% is cost of labeling data via humans is is defined by  is usually a proxy for the budget $b$

% todo...

% \section{LLMs for Active Learning vs. Active learning for LLMs \writer{Yu Xia}}
%\ryan{%
%Active learning for LLMs vs. LLM for Active Learning
%}

\section{AL Paradigms with LLMs \writer{Yu Xia}} \label{sec:setting}
% \ryan{move}

With the rise of LLMs, AL has evolved to address new challenges and opportunities across various learning paradigms. In this section, we briefly outline four LLM-based AL paradigms, where more details can be found at Appendix \ref{app:setting}. 
We also take a step further and briefly discuss in Appendix \ref{sec:unifying-settings} a unifying view of the LLM-based AL problem through the lens of bandits, online learning, RLHF, pandoras box, among others.

\paragraph{Active In-Context Learning} Recent studies frame few-shot demonstration selection as an active learning problem, leveraging semantic coverage and ambiguity-driven sampling to optimize LLM performance~\citep{margatina2023active, mavromatis2024covericl, qian2024ape}. 

\paragraph{Active Supervised Fine-Tuning} To reduce labeling costs, active learning has been integrated into supervised fine-tuning via uncertainty-based querying, self-training on low-uncertainty data, and strategic sample selection~\citep{yu-etal-2022-actune, xia-etal-2024-hallucination, bayer2024activellm}. 

\paragraph{Active Preference Alignment} Efficient label selection is critical in reinforcement learning from human feedback (RLHF). Recent approaches employ targeted preference queries to accelerate alignment and improve data efficiency~\citep{ji2024reinforcement, muldrew2024active, chen2024cost}.

\paragraph{Active Knowledge Distillation} Selective knowledge transfer from LLMs to smaller models reduces computational costs. Methods using uncertainty-based sample selection and iterative student feedback improve distillation efficiency while maintaining performance~\citep{zhang2024elad, liu-etal-2024-evolving, palo-etal-2024-performance}.

\section{Applications \writer{Ryan Aponte}} \label{sec:apps}

LLM-based active learning (AL) has been applied across diverse tasks, reducing annotation costs and improving performance in data-scarce settings. 
We summarize in Table~\ref{tab:applications} key applications and specific use cases.
Table~\ref{table:qual-and-quant-comparison} also bridges these applications with our taxonomy of techniques, which provides an intuitive comparison of the state-of-the-art LLM-based AL methods.
These applications include text classification~\citep{rouzegar2024enhancing}, text summarization~\citep{li2024active}, non-text classification~\citep{margatina2023active}, question answering~\citep{diao2023active}, entity matching~\citep{qian2024ape}, debiasing~\citep{du2024causal}, translation~\citep{kholodna2024llms}, and sentiment analysis~\citep{xiao2023freeal}. Beyond these, AL has also been used for optimizing system design~\citep{taneja2024can} and question generation~\citep{piriyakulkij2023active}. 
For a more detailed discussion of the AL applications on each task, please refer to Appendix~\ref{app:apps}, where we also include a pairing of active learning applications and datasets in Table~\ref{tab:AL-datasets}.
In Appendix~\ref{sec:appendix-AL-objectives}, we also discuss in further detail the benefits of applying LLM-based AL.

\section{Open Problems \& Challenges \writer{Zhouhang Xie}} \label{sec:open-problems-challenges}

In this section, we discuss open problems and challenges of LLM-based AL for future works.

\paragraph{Heterogeneous Annotation Costs} Traditional AL assumes a fixed annotation budget, but LLM-based AL introduces complex cost structures, including human labeling, LLM query costs, and annotation expenses. Future work should develop algorithms that optimize selection strategies while accounting for these heterogeneous costs.

\paragraph{Multi-LLM AL Algorithms} Different LLMs present trade-offs in performance and cost, suggesting opportunities for hybrid approaches that combine weak-cheap and strong-expensive models. Inspired by retrieval-augmented models~\citep{huang2024comprehensivesurveyretrievalmethods} and multi-oracle clustering~\citep{DBLP:conf/iclr/SilwalANMRK23}, optimizing multi-LLM frameworks for AL remains an open challenge.

\paragraph{LLM Agents and Active Learning} Integrating AL into LLM agents presents new possibilities, such as improving retrieval-augmented generation (RAG)~\citep{xu2024activeragautonomouslyknowledgeassimilation} and in-context example selection~\citep{mukherjee2024experimental}. Conversely, exploring LLM agents as annotation tools could enhance existing AL pipelines by reducing human labeling costs.

\paragraph{Multimodal Active Learning} Most LLM-based AL applications focus on text, leaving open questions on extending these methods to images, audio, and behavioral data. Future work should investigate how well LLMs generalize to non-text domains and whether they can match human-level annotation quality in these settings.

\paragraph{Complex Feedback and Multi-Aspect Labels} Traditional AL usually relies on single-label supervision, but reward models in the era of LLMs are able to return structured feedback that scores multiple facets of a response \cite{bai2022training}, such as factual accuracy, completeness, and fluency. Developing LLM-based AL strategies that can leverage these rich signals without overfitting to any one dimension may be a promising solution.

\paragraph{Unstable Performance in LLM-based Annotation} While LLMs are increasingly used to simulate human annotations~\citep{zheng2023judgingllmasajudgemtbenchchatbot, lambert2024rewardbenchevaluatingrewardmodels}, their reliability varies across domains~\citep{tan2024largelanguagemodelsdata}. Future research should explore adaptive routing mechanisms that dynamically allocate annotation tasks between LLMs and human annotators based on model competency.

%\todob{How about complex feedback? For instance, reward models in NLP typically judge many aspects of the text, such as accuracy, completeness, and fluency.}

%\subsection{Multi-objective Active Learning}

%\subsection{Hierarchical Active Learning}

%\subsection{Active Multi-LLM and Multi-Agent Techniques}

% \subsection{Theoretical Foundations}
% We need to develop better theoretical foundations for LLM+AL, especially when a black-box oracle is involved...

\section{Conclusion \writer{TODO}} \label{sec:conc}
% \todo{}
In this survey, we present an intuitive taxonomy of LLM-based Active Learning, detailing how LLMs can act as sample selectors, data generators, and annotators within the AL loop. 
We show how these techniques are reshaping traditional AL paradigms and enabling more efficient data acquisition and model training across various applications.
By reviewing existing methods, highlighting emerging trends and discussing open challenges, we aim to offer a useful foundation for researchers and practitioners looking to incorporate LLM-based AL techniques into their applications.

\section*{Limitations}
% Our work has several limitations.
% 
%In this paper, we do not summarize datasets as most approaches can leverage a wide variety of datasets.
%Many examples provided throughout the paper are based on classification for simplicity, though LLM-based AL techniques are more widely applicable for other tasks.
% First, ...
% Second, ...
% Third, ...
% These are left to future work.
% \todo{please improve}
LLM-based Active Learning (AL) gives rise to many important applications with critical advantages over traditional AL techniques. Despite the fundamental importance of LLM-based AL, there remain some challenges. The reliance on high-quality initial labeled data may introduce biases, while the computational overhead, cost, and robustness of iterative query generation and selection may limit its application in practice. Query generation and selection techniques via LLMs remain sensitive to model uncertainty, often lacking theoretical guarantees and may  lead to inconsistent performance. Furthermore, such techniques may also suffer from reproducibility and robustness issues. Ethical risks such as bias amplification through generation of examples and labels outside known set remains important to handle when deployed in practice.

\bibliography{main}
\bibliographystyle{acl_natbib}

% \newpage
\appendix
\newpage
\section{LLM-based AL Goals}\label{app:al_setting}

In this section, we highlight the various ways LLM-based AL can be used by providing an intuitive categorization based on the LLM-based AL goals.
In particular, LLM-based AL techniques can focus on 
selecting or generating specific data points $\mX$,
learning an optimal prompt $P$,
selecting contexts $C$ to include in the prompt,
deciding on the LLM model $\mathcal{M}_i \in M$,
features to leverage $x_j$,
hyperparameters $\theta$,
model architectures $\mathcal{A}$,
data for annotation $\mX_{S}$,
and
evaluation data $\mX_{\rm eval}$.

\begin{itemize}[left=0pt]
    \item \textbf{Data Points} ($X$): 
    LLM-based active learning can be used to select specific data points $x \in X$ from a larger pool for different purposes, such as training, budget-constrained training, and domain adaptation. This also includes selecting data for annotation, $x_{\text{annot}}$, which maximizes learning gains, and choosing evaluation subsets, $x_{\text{eval}}$, that ensure comprehensive model assessment.
    Furthermore, we can also leverage LLMs to generate entirely novel data points that lie outside the set of unlabeled data.
    This 
    
    \item \textbf{Prompts} ($P$): 
    The technique can be applied to select the most effective prompts $p \in P$ and prompt variations that optimize LLM outputs for specific tasks.
    
    \item \textbf{Contexts} ($C$): 
    LLM-based active learning can be utilized to select the most relevant contexts $c \in C$ or contextual inputs that enhance model performance on context-dependent tasks.
    
    \item \textbf{LLM Model Variants} ($\mathcal{M}$): 
    The method can be used to select the best LLM model variant $\mathcal{M}_i \in \mathcal{M}$ from a set of available models for a given input.
    Alternatively, we can also use LLM-based AL to identify models that contribute effectively to ensembles.
    
    \item \textbf{Features} ($x_j$): 
    LLM-based active learning can be used to select important features $x_j \in \vx$ or identify new features that can be integrated to improve model accuracy and explainability.
    Furthermore, it can also be leveraged to estimate missing features, e.g., if there are several missing values in the feature vector of a specific instance.
    
    \item \textbf{Hyperparameters} ($\theta$): 
    The technique is useful for selecting optimal hyperparameter values $\theta_i \in \theta$, such as learning rate or batch size, and adjusting them dynamically during training to optimize performance.
    
    \item \textbf{Model Architectures} ($\mathcal{A}$): 
    LLM-based active learning can assist in selecting the most appropriate model architecture $\mathcal{A}_i \in \mathcal{A}$ or choosing between different versions of a model for specific use cases.
    
    \item \textbf{Data for Annotation} ($\mX_{S}$): 
    The technique can be used to select specific data subsets $\mX_{S} \subset \mX$ that should be labeled by human annotators, focusing on those that provide the greatest potential improvement in model learning.
    
    \item \textbf{Evaluation Data Subsets} ($\mX_{\text{eval}}$): 
    LLM-based active learning is beneficial for selecting evaluation data points $\mX_{\text{eval}} \subset X$ that maximize the effectiveness of model evaluation, ensuring coverage of edge cases and comprehensive testing.
\end{itemize}
\section{Traditional Selection Strategies}
\label{app:trad-strategy}
We provide additional discussions on traditional selection strategies from Section~\ref{sec:querying-selection}.
The \text{CoreSet} is a pure diversity-based strategy where unlabeled examples are selected using a greedy furthest-first traversal conditioned on all labeled examples \citep{sener2017active, geifman2017deep, citovsky2021batch}. Similarly \citet{agarwal2020contextual} uses a coreset-based strategy on features to select unlabeled examples using CNN. The \text{Least confidence} is an uncertainty-based active learning algorithm where the uncertainty score of an unlabeled example is its predicted class probability and the algorithm then samples unlabeled examples with the smallest uncertainty score \citep{settles2009active, settles2011theories, wang2014new}. The \text{Margin}-based selection strategy is also an uncertainty-based strategy \citep{tong2001support, balcan2009agnostic, settles2009active}. Margin first sorts unlabeled examples according to their multiclass margin score and then selects examples that are the hardest to discriminate and can be thought of as examples closest to their class margin. The \text{Max-entropy} strategy \cite{wang2014new, kremer2014active, diao2023active} is an uncertainty-based strategy that selects unlabeled examples according to the entropy of the example's predictive class probability distribution. The \text{BADGE} algorithm combines both uncertainty and diversity sampling \citep{ash2019deep, ash2021gone}. Badge chooses a batch of unlabeled examples by applying $k$-Means++ \citep{arthur2006k} on the gradient embeddings computed from the penultimate layer of the model. The value of the gradient embedding captures the uncertainty score of the examples. Finally, \text{BALD} (Bayesian Active Learning by Disagreements)  \citep{kirsch2019batchbald, pmlr-v70-gal17a} chooses unlabeled examples that are expected to maximize the information gained from the model parameters i.e. the mutual information between predictions and model posterior.

% \subsection{Budget}
% % one facet of the problem
% In LLM-based AL, one may also have a fixed budget on the cost of generation (\eg, input/output tokens).

% Budget:
% - Number of samples
% - LLM cost of generation (input/output tokens)
% - etc

% % % \todo{revise text}
% % Budget in terms of the monetary cost of generation via LLMs:
% % Note that there are two costly elements, e.g., we may have a budget for the cost of obtaining the output $Y_{ij}$ for a given model $M_i \in \mathcal{M}$ and input prompt $X_j \in \mathcal{X}$.
% % This is difficult as the cost is based on the number of input tokens (which can be computed apriori, and the number of tokens of the output from the model, which is unknown, but can be somewhat estimated or upper bounded).

% % Alternatively, the problem may be formulated with respect to a budget $B$ on the human feedback for a specific model $M_i$, input prompt $X_j$, and output $Y_{ij} = M_i(X_j)$, forming a tuple $(M_i, X_j, Y_{ij}=M_i(X_j))$.

% \subsection{Add other facets}
% Discuss other unique facets to this problem, key differences, new challenges, and opportunities.

\eat{%}
\todo{%}
In work by~\citet{atsidakou2024contextual} it is shown that contextual Pandoras box can be reduced to contextual bandits and further reduced to online regression.
}%
\todo{cite other works, etc.}
% From Contextual Bandits to bandits to online Regression

\todo{Add unification and show that these are recovered (certain instantiations)}

\ryan{cite relevant papers for known connections, bandits--pandoras, contextual bandits online linear regression,} 
}%

\section{AL Paradigms with LLMs \writer{Yu Xia}} \label{app:setting}
%\ryan{%
%Active learning for LLMs vs. LLM for Active Learning
%}%

%\subsection{Active Learning for LLMs}
With the rise of LLMs, the active learning has evolved to address new challenges and opportunities across various learning paradigms. 
%In the following sections, we discuss several key areas where active learning is reimagined for LLM-centric workflows.

%\subsection{Active Prompt Engineering}
%Traditional prompt design, especially when incorporating few-shot examples, requires extensive manual effort to identify most informative demonstrations.
%To address this, \citet{qian2024ape} propose APE (Active Prompt Engineering), a human-in-the-loop tool that actively selects ambiguous examples for human feedback, which are then converted into few-shot exemplars to improve LLM performance.
%To further reduce the human effort, \citet{bayer2024activellm} introduce ActiveLLM, an approach leveraging directly LLMs for selecting informative examples in textual few-shot scenarios.

%There have also been work using AL techniques for finding the best few-shot examples.
%\citet{qian2024ape} develops an active learning approach for prompt engineering where at each iteration, a set of prompts are derived, and then evaluated by a committee of models, where the best is selected, and then the approach repeats.
%\cite{bayer2024activellm}

\subsection{Active In-Context Learning}
Recent advances have recast demonstration selection for in-context learning as an active learning problem, where the goal is to identify and annotate the most informative examples under stringent labeling budgets. For example, \citet{margatina2023active} demonstrate that similarity-based sampling can consistently outperform traditional uncertainty-based methods when selecting single-round demonstrations for LLMs. Building on these insights, \citet{mavromatis2024covericl} propose CoverICL—a graph-based algorithm that integrates uncertainty sampling with semantic coverage to enhance both performance and budget efficiency across diverse tasks and LLM architectures. In a complementary approach, \citet{qian2024ape} introduce APE, a human-in-the-loop tool that iteratively pinpoints ambiguous examples for few-shot prompts, thus progressively refining LLM performance through active learning principles.

\subsection{Active Supervised Finetuning}
Active learning strategies have also been adapted to reduce the labeling cost associated with supervised finetuning. For instance, \citet{yu-etal-2022-actune} present AcTune, which actively queries high-uncertainty instances while simultaneously leveraging self-training on low-uncertainty unlabeled data. This dual strategy, further refined by region-aware sampling, effectively mitigates redundancy in the training data. Similarly, \citet{xia-etal-2024-hallucination} propose an active learning framework tailored for text summarization that systematically identifies and annotates diverse instances exhibiting various types of hallucinations, thereby improving both data efficiency and factual correctness. Extending these ideas to overcome cold-start challenges, \citet{bayer2024activellm} develop ActiveLLM, which employs state-of-the-art LLMs (e.g., GPT-4, Llama 3) to select informative instances, significantly boosting finetuning performance in both few-shot and iterative settings.

\subsection{Active Preference Alignment}\label{app:rlhf}
Active preference alignment targets label-efficient methods for refining LLMs via human or AI feedback.
\citet{ji2024reinforcement} frame reinforcement learning from human feedback (RLHF) as a contextual dueling bandit problem and develop an active-query-based algorithm (ADPO) that drastically cuts down the number of preference queries needed for LLM alignment.
Building on simpler, more stable methods, \citet{muldrew2024active} propose an active learning extension to Direct Preference Optimization (DPO), showing notable gains in both convergence speed and final quality through selective preference labeling.
Complementarily, \citet{chen2024cost} introduce a cost-effective approach for constructing reward models, combining on-policy querying and active data selection to maximize the impact of limited human feedback, and achieving strong performance improvements in DPO with minimal expert annotation.

\subsection{Active Knowledge Distillation}
Active knowledge distillation has emerged to reduce the computational costs of LLMs by selectively transferring their knowledge into smaller models. 
\citet{zhang2024elad} propose ELAD, which leverages reasoning-step uncertainties to guide sample selection and employs teacher-driven explanation revisions to optimize the distillation process. 
\citet{liu-etal-2024-evolving} introduce EvoKD, an iterative strategy that identifies student model weaknesses and dynamically generates labeled data, continuously refining the student’s capabilities through LLM feedback. 
Meanwhile, \citet{palo-etal-2024-performance} present PGKD, an active feedback loop that uses performance signals such as hard-negative mining to inform new data creation, yielding substantial efficiency gains and reducing inference costs for text classification at scale.

% \begin{comment}
% \subsection{Unifying Framework: From AL to Bandits and Beyond}
\section{From AL to Bandits and Beyond}
% : A Unification}
\label{sec:unifying-settings}
%, Online Learning, RLHF, Pandoras Box, and Beyond}
% \ryan{add table sketched in notebook}

%Also fine-tuning, unlearning, dynamical systems etc
In this section, we briefly discuss a unifying view of the LLM-based AL problem through the lens of bandits, online learning, RLHF, pandoras box, among others.
We provide an intuitive summary of such connections in Table~\ref{tab:comparison-problem-settings} across a variety of problem settings.
%We also provide a more fine-grained analysis in Table~\ref{tab:AL-connections-overview}.
Intuitively, they overlap in goals, that is, 
% Overlap in Goals: 
many of these techniques aim to optimize resource use (e.g., labels, feedback, compute) while improving learning. 
They nearly all operate sequentially, but they differ in what drives decisions (e.g., reward, uncertainty, cost).
% Thus, they are all sequent
% 
However, they may have different constraints, for instance, 
% Differences in constraints: 
some settings assume explicit costs (e.g., pandora's box), while others lack such constraints (e.g., online learning).
% 
% Sequential Learning: Nearly all approaches operate sequentially, but they differ in what drives decisions (e.g., reward, uncertainty, cost).
These insights allow practitioners to choose the right framework depending on the data, task, and constraints of their problem.

\renewcommand{\arraystretch}{1.5} % Increase row height for readability

\begin{table}[t!]
\centering
\footnotesize
% \rowcolors{2}{gray!10}{white}
\begin{tabular}{p{2.5cm} *{4}{>{\centering\arraybackslash}m{0.3cm}}
H
H
*{2}{>{\centering\arraybackslash}m{0.5cm}}
}
\toprule
% \textbf{Technique} 
& 
\rotatebox{90}{\textbf{Sequential Learning}} & 
\rotatebox{90}{\textbf{Active Querying}} & 
\rotatebox{90}{\textbf{Human Feedback}} & 
\rotatebox{90}{\textbf{Exploration-Exploitation}} & 
\rotatebox{90}{\textbf{Cost-Awareness}} & 
\rotatebox{90}{\textbf{Pre-trained Models}} & 
\rotatebox{90}{\textbf{Statistical Foundations}} & 
\rotatebox{90}{\textbf{Dynamic Environments}} \\
\midrule

\textbf{Active Learning} & 
\cmark & \cmark &  & \cmark &  &  & \cmark & \cmark \\

\textbf{RLHF} & 
\cmark & \cmark & \cmark & \cmark &  &  &  & \cmark \\

\textbf{Pandora’s Box} & 
\cmark & \cmark &  & \cmark & \cmark &  & \cmark & \cmark \\

\textbf{Bandits} & 
\cmark &  &  & \cmark &  &  &  & \cmark \\

\textbf{Optimal Design} & 
 &  &  &  & \cmark &  & \cmark &  \\

\textbf{Online Learning} & 
\cmark &  &  & \cmark &  &  & \cmark & \cmark \\

% \textbf{Fine-tuning} & 
%  &  &  &  &  & \cmark &  & \cmark \\

\bottomrule
\end{tabular}

\caption{%
%\ryan{this needs to be carefully checked, for instance, all use human feedback right (under the def we have below)? Also some other factors seem to be missing, and others may not be that meaningful/useful.}
Comparison of problem settings across a range of important properties.
\textit{Sequential Learning}: Learning incrementally over time, rather than all at once. 
\textit{Active Querying}: Actively selecting or querying data points to improve the model. 
\textit{Human Feedback}: Utilizing human input or feedback to guide and improve learning. 
\textit{Exploration-Exploitation}: Balancing the need to explore new options versus exploiting known ones. 
\textit{Cost-Awareness}: Considering the costs of acquiring data or feedback during the learning process. 
% \textbf{Pre-trained Models}: Using pre-existing models as a starting point for further learning or adaptation. 
\textit{Statistical Foundations}: Grounding the technique in statistical principles or experimental design. 
\textit{Dynamic Environments}: Adapting to environments that change over time or where new information continuously emerges.
% \eat{%}
% \todo{Fix, add more factors, add grey checkmark, and define the properties more formally/intuitively that are used to summarize, etc}
% \todo{Still need to fix some stuff and improve. Should add math formulation, etc.}
% }%
}
\label{tab:comparison-problem-settings}
\end{table}

\section{Applications \writer{Ryan Aponte}} \label{app:apps}
LLM-based active learning offers many promising applications; from LLM-to-SLM knowledge distillation  to low-resource language translation, to entity matching. Active learning reduces the cost of data supervision, broadening the tasks machine learning can be applied to. Active learning, combined with causal learning, may also be useful for reducing measured bias (Sec.~\ref{apps-active-debiasing}).
Table~\ref{tab:taxonomy-techniques} offers a taxonomy of active learning techniques. 
Overall, active learning is likely to continue to offer benefits in data-constrained environments. 
In particular, active learning may increase the utility of LLMs in domains requiring expert knowledge, such as medicine, law, and engineering; a historical weakness~\cite{yao2023beyond}. Finally, Table~\ref{tab:AL-datasets}, we include a pairing of active learning applications and datasets used.

% \ryan{I added initial text (currently a big larger than a page, but much to do) and created an initial application table (which needs updating to be consistent with the subsections, and we can add Sec.~\ref{} to every application subsection in Table}

% \ryan{Ryan Aponte: Can you polish the table, add to it, add Sec, and make it consistent with subsections, then add any other subsections I missed with work, and then add intro paragraph to this section, and final discussion subsection that discusses other applications, or novel aspects that arise.}

% \ryan{Ryan Aponte: when you are going through the works, please add the datasets used in various works, the application, etc. Please create a new table for this, as it will help with discussion, and can be used later for datasets section too}

\subsection{Text Classification}\label{apps-text-classification}

\citet{rouzegar2024enhancing} apply active learning to identify the most relevant samples for labeling text. The framework is applied to, among other datasets, Fake News for document authenticity~\cite{avrv-tp46-24}. The method reduces the cost of obtaining supervision and also enables a trade-off between cost efficiency and performance. The authors use GPT-3.5 and require some human supervision, unlike some more recent work~\cite{du2024causal}. 
ActiveLLM~\cite{bayer2024activellm} uses LLMs for selecting instances for few-shot text classification with model mismatch, in which the selection, or query, model is different from the model used for the final task. ActiveLLM use an LLM to estimate data point uncertainty and diversity without external supervision, improving few-shot learning.

\subsection{Text Summarization}\label{apps-text-summarization}

\citet{li2024active} proposes LLM-Determined Curriculum Active Learning (LDCAL) that improves the stability of the active learner by selecting instances from easy to hard by using LLMs to determine the difficulty of a document. LDCAL also leverages a new AL technique termed Certainty Gain Maximization that captures how well the unselected instances are represented by the selected ones, which is then used to select instances that maximize the certainty gain for the unselected ones.
More formally, the certainty gain (CG) measures the gain in representation certainty for an unlabeled instance $\vx_u$ when a candidate instance $\vx_s$ is selected for annotation, that is,
\begin{align}
\text{CG}(\vx_{s}, \vx_{u}) \!= \!\max \!\left(\! f(\vx_{s}, \vx_u) \!- \!\max_{\vx_{i} \in \mathcal{D}_{\ell}} \!f(\vx_{u}, \vx_{i}), 0 \!\right)\nonumber
\end{align}
It is derived as the maximum of the difference between $f(\vx_s, \vx_u)$, which measures the similarity between $\vx_s$ and $\vx_u$, and the current maximum similarity $\max_{\vx_i \in \mathcal{D}_\ell} f(\vx_u, \vx_i)$, where $\vx_i$ represents instances in the labeled set $\mathcal{D}_\ell$. To ensure non-negative values, any negative difference is clipped to 0. This formulation ensures that the selection of $\vx_s$ positively contributes to the representational coverage of the unlabeled instances, balancing the sampling process across high-density and low-density regions in the data distribution.
Finally, the Average Certainty Gain (ACG) for a candidate instance $\vx_s$ is
\[
\text{ACG}(\vx_s) = \frac{1}{L} \sum_{\vx_u \in \mathcal{D}_{\text{u}}} \text{CG}(\vx_s, \vx_u)
\]
where $L$ is the total number of unlabeled instances and $\mathcal{D}_{\text{u}}$ is the set of all unlabeled instances.
A key advantage of LDCAL over uncertainty-based acquisition strategies is that the acquisition model does not need to be trained after each AL iteration.

\subsection{Non-Textual Classification}\label{apps-non-text-classification}
~\citet{margatina2023active} apply active learning to several areas, including multiple-choice question answering and test on 15 different models from the GPT and OPT families. The selection of similar in-context samples for multiple-choice questions was the most effective sampling criterion. For classification, diversity was more effective than similarity as selection criterion. The authors also found larger models to have higher performance.

\subsection{Question Answering}\label{apps-question-answering}
\citet{diao2023active}, in recognition of the utility of chain-of-thought prompting, ActivePrompt uses active learning to design more effective prompts for LLMs based on human-designed chains of thought. With only eight exemplars made by humans, the method achieves higher performance on complex reasoning tasks. Uncertainty is estimated by querying an LLM with the same prompt repeatedly to and response disagreement is measured.

\subsection{Entity Matching}\label{apps-entity-matching}

An approach called APE~\cite{qian2024ape} focuses on an active prompt engineering approach for entity matching where at each iteration, a set of prompts are derived, and then evaluated by a committee of models, where the best is selected, and then the approach repeats. APE selects the most informative samples of a dataset reduces the cost to humans of identifying samples in most need of human feedback. 
%In a demo video, DBLP-Scholar is sampled from~\cite{}.
% 
\citet{ming2024autolabel} proposed AutoLabel that starts by selecting the most representative seed data using traditional techniques such as density clustering and sampling. Then uses an LLM with chain-of-thought prompting to obtain labels, and then leverages human feedback to rectify the labeled results for entity recognition.
\citet{zhang2023llmaaa} proposed LLMaAA that leverages LLMs for annotation in an active learning loop. LLMaAA is used for both named entity recognition and relation extraction.

\begin{table}[!t]
\centering
\small
\setlength{\tabcolsep}{0pt} % Reduce column separation
\begin{tabular}{%
    p{0.45\linewidth}
    p{0.45\linewidth}
}
\toprule
\textbf{Task} & \textbf{Datasets} \\
\midrule
\textbf{Text Classification} & IMDB~\cite{maas-EtAl:2011:ACL-HLT2011} \\
\textbf{Text Summarization} & AESLC~\cite{zhang-tetreault-2019-email} \\
\textbf{Non-Text Classification} & Crossfit~\cite{ye2021crossfitfewshotlearningchallenge} \\
\textbf{Question Answering} & Crossfit~\cite{ye2021crossfitfewshotlearningchallenge} \\
\textbf{Entity Modeling} & DBLP-Scholar~\cite{dblpscholar} \\
\textbf{Debiasing} & BBQ~\cite{parrish2022bbqhandbuiltbiasbenchmark}, MT-Bench~\cite{zheng-etal-2023-judging}, UNQOVER~\cite{li-etal-2020-unqovering} \\
\textbf{Translation} & IT domain~\cite{koehn-knowles-2017-six} \\
\textbf{Sentiment Analysis} & IMDB~\cite{maas-EtAl:2011:ACL-HLT2011}, AESLC~\cite{zhang-tetreault-2019-email}, AGNews~\cite{NIPS2015_250cf8b5} \\
\bottomrule
\end{tabular}
\caption{LLM-based Active Learning Datasets}
\label{tab:AL-datasets}
\end{table}

\subsection{Active Debiasing of LLMs}\label{apps-active-debiasing}

\citet{du2024causal} proposed a causal-guided AL approach for debiasing LLMs by leveraging the LLMs to select data samples that contribute bias to the dataset. The method works by applying active learning to identify the most important samples within the dataset and the model looks for causally invariant relationships; this approach is less compute-intensive than fine-tuning a model on a debiasing dataset.

\subsection{Translation}\label{apps-translation}

\citet{kholodna2024llms} apply active learning for annotations in low-resource languages and find near-SOTA performance, with a reduction in annotation cost (relative to human annotators) of 42.45 times. Like other active learning methods, samples with the highest prediction uncertainty are selected. The authors test on 20 low-resource languages spoken in Sub-Saharan Africa. The authors found larger LLMs like GPT-4-Turbot and Claude 3 Opus had more consistent performance than Llama 2-70B and Mistral 7B. 
FreeAL use active learning to collect data for task-specific knowledge, such as translation~\cite{xiao2023freeal}, without requiring human annotation. Tested on eight benchmarks, FreeAL achieves near-human-supervised performance, without requiring any human annotation. With additional feedback rounds, FreeAL was able to improve performance. In FreeAL, an LLM and SLM are paired and the LLM provides annotation, while the SLM is a weak learner. The authors propose the use of limited human supervision to further improve performance.

\subsection{Sentiment Analysis}\label{apps-sentiment-analysis}
~\citet{xiao2023freeal} use active learning for movie sentiment analysis with the Movie Review dataset Seeing Stars~\cite{pang2005seeingstarsexploitingclass} in the method ActivePrune, which uses ordinal (movie stars, 1-5) labels.
ActivePrune outperforms other pruning methods and with its increased compute efficiency, they reduce end-to-end active learning time by 74\%. ActivePrune works by reducing dataset size and increasing representation of underrepresented data, based on perplexity.

\subsection{Other}\label{sec-apps-other}
% \todo{move to appropriate place in paper}
% RyanA - there is an "other" section in the table, so one is included here
In this section, we describe additional noteworthy applications of active learning.

\subsubsection{Question Generation}\label{sec-apps-other-question-generation}
\citet{piriyakulkij2023active} proposed an active preference inference approach that infers the preferences of individual users by minimizing the amount of questions to ask the user to obtain their preferences.
The approach uses more informative questions to improve the user experience of such systems.

\subsubsection{System Design}\label{sec-apps-other-system-design}
\citet{taneja2024can} leverages an approach for actively correcting labels to enhance LLM-based systems that have multiple components.
\citet{astorga2024partially} introduced a partially observable cost-aware AL method focused on the setting where features and/or labels may be partially observed.

\section{LLM-based AL Advantages}\label{sec:appendix-AL-objectives}
By iteratively selecting and generating instances (to label) for training, LLM-based active learning can have the following advantages:

\begin{itemize}[left=0pt]
    \item \textbf{Better Accuracy:} LLM-based active learning can achieve better accuracy with fewer instances by selecting and generating the most informative instances.

    \item \textbf{Reduced Annotation Costs:} LLM-based active learning techniques can reduce labeling and other annotation costs by selecting or generating the most informative samples to use for model training and fine-tuning.

    \item \textbf{Faster Convergence:} Using LLM-based active learning often leads to faster convergence as the model can be learned more quickly by iteratively selecting and generating the most informative examples.

    \item \textbf{Improved Generalization:} By leveraging LLM-based active learning techniques that iteratively select and generate the most informative and diverse examples, the model can generalize better to new data.

    \item \textbf{Robustness:} Iteratively selecting or generating the best examples for training can often improve the robustness to noise, as the model is learned from a set of high-quality representative examples.
\end{itemize}

\end{document}